\documentclass[10pt]{article}
\usepackage[english]{babel}
\usepackage{fullpage}
\usepackage{geometry}
\usepackage{graphicx}
\usepackage{amsmath}
\usepackage{amsthm,amssymb,amsfonts,mathtools}
\usepackage{marvosym}
\usepackage{hyperref}
\usepackage{color}
\usepackage{fontawesome}
\usepackage{pdfpages}
\usepackage{caption}
\usepackage{algorithm,algpseudocode}
\usepackage{caption}
\usepackage{fdsymbol}
\usepackage{float}
\usepackage{authblk}

\usepackage{url,hyperref,lineno,microtype,subcaption}
\usepackage[onehalfspacing]{setspace}

\newcommand{\D}{\mathrm{d}}
\newcommand{\Pa}{\mathrm{\partial}}

\makeatletter
\renewcommand{\maketitle}{\bgroup\setlength{\parindent}{0pt}
\begin{flushleft}
    {\Large \textbf{\@title}}

  \@author
\end{flushleft}\egroup
}
\makeatother

\usepackage{abstract}

\title{Neural and Synaptic Array Transceiver: 
        A Brain-Inspired Computing Framework for Embedded Learning} 

\author[1]{Georgios Detorakis}
\author[4]{Sadique Sheik}
\author[2]{Charles Augustine}
\author[2]{Somnath Paul}
\author[3]{Bruno U. Pedroni}
\author[5,1]{Nikil Dutt}
\author[1,5]{Jeffrey Krichmar}
\author[3]{Gert Cauwenberghs}
\author[1,5]{Emre Neftci}
\date{}

\affil[1]{Department of Cognitive Sciences, UC Irvine, Irvine, CA, USA 92697}
\affil[2]{Intel Corporation - Circuit Research Lab, Hillsboro, OR, USA 97124}
\affil[3]{Department of Bioengineering and Institute for Neural Computation, UC San Diego, La Jolla, CA, USA 92093}
\affil[4]{Biocircuits Institute, UC San Diego, La Jolla, CA, USA 92093}
\affil[5]{Department of Computer Science, UC Irvine, Irvine, CA, USA 92697}

\begin{document}

\maketitle

\begin{flushleft}
\begin{abstract}
    \noindent
Embedded, continual learning for autonomous and adaptive behavior is a key application of neuromorphic hardware.
However, neuromorphic implementations of embedded learning at large scales that are both flexible and efficient
have been hindered by a lack of a suitable algorithmic framework. 
As a result, most neuromorphic hardware are trained off-line on large clusters of dedicated processors or GPUs
and transferred \emph{post hoc} to the device. 
We address this by introducing the neural and synaptic array transceiver (NSAT), a neuromorphic computational framework facilitating flexible and efficient embedded learning by matching algorithmic requirements and neural and synaptic dynamics. 
NSAT supports event-driven supervised, unsupervised and reinforcement learning algorithms including deep learning.  
We demonstrate the NSAT in a wide range of tasks, including the simulation of Mihalas--Niebur neuron, dynamic
neural fields, event-driven random back-propagation for event-based deep learning, event-based contrastive 
divergence for unsupervised learning, and voltage-based learning rules for sequence learning. 
We anticipate that this contribution will establish the foundation for a new generation of devices enabling 
adaptive mobile systems, wearable devices, and robots with data-driven
autonomy. 
\end{abstract}
\end{flushleft}

\section{Introduction}

Brain-inspired computing paradigms can lead to massively distributed technologies that compute on extremely tight power budgets, while being robust to ambiguities in real-world sensory information and component failures.
To devise such technology, \emph{neuromorphic} electronic systems strive to mimic key building blocks of biological neural networks and dynamics \cite{Mead89_analvlsi} in custom digital \cite{Merolla_etal14_millspik,Furber_etal14_spinproj} or mixed signal \cite{Schemmel_etal10_wafeneur,Benjamin_etal14_neurmixe,Qiao_etal15_recoon-l} CMOS technologies.



Recent progress has significantly advanced the systematic synthesis of dynamical systems onto neural substrates and their neuromorphic counterparts. 
For instance, the configuration of spiking neural networks for inference tasks have been solved using frameworks such as the neural engineering framework \cite{Eliasmith_Anderson04_neurengi}  and STICK \cite{Lagorce_Benosman15_sticspik}, direct mapping of pre-designed \cite{Neftci_etal13_syntcogn} or pre-trained neural networks \cite{Cao_etal14_spikdeep}. 

Many of these approaches were successfully ported to neuromorphic hardware \cite{Qiao_etal15_recoon-l,Esser_etal16_convnetw,Neftci_etal13_syntcogn}.
While these solutions are promising from an energetic point of view in inference tasks, they heavily rely on computers (GPU or CPU) for their configuration and largely abandon adaptive and autonomous behavior capabilities in the presence of intrinsic and extrinsic variations. 
These critical features can be introduced through embedded synaptic plasticity and learning ``on-the-fly''.
However, learning using data streaming to the neuromorphic device presents significant challenges. 
One challenge is technological: synaptic plasticity requires high memory bandwidth, but the realization of adequate high density memory co-located with the neuron is costly using current technologies. 
While emerging memory technologies are poised to solve this problem, the solutions remain difficult to control and lack precision. 
Another challenge is algorithmic: the co-location of memory with the neuron leads to significant algorithmic challenge: state-of-the-art algorithms in machine learning rely on information that is temporally and spatially global when implemented on a neural substrate. 
Finally, the hardware implementation of learning involves a hard commitment to the plasticity dynamics, but doing so in a way that is both hardware-friendly and capable of learning a wide range of tasks is a significant modeling challenge.
Our recent work in neuromorphic algorithms demonstrated that most algorithmic
challenges can be solved
\cite{Eliasmith_etal12_largmode,Lagorce_Benosman15_sticspik}, and can
potentially result in learning systems that require a thousandfold less power
than mainstream
technologies~\cite{Neftci_etal16_stocsyna,Neftci_etal17_evenranda,Mostafa_etal17},
while matching or surpassing the accuracy of dedicated machine learning
accelerators, and operating on-line. \textcolor{black}{In addition, neuromorphic learning-enabled devices are expected to have similar
energy per operation figures with learning-enabled artificial neural networks, such 
as binary neural networks~\cite{Neftci_2018_data}. Furthermore, it has been shown 
that neural networks with binary activations are a class of spiking neural networks (without states or dynamics)~\cite{Neftci_2018_data}, implying that the proposed framework is capable of implementing neural networks without binary activations as well.}

One outstanding question is whether 
one can formulate a general event-based learning rule that is general and capable of learning a wide range of tasks
while being efficiently realizable using existing memory technologies.

This article presents one such system, called Neural and Synaptic Array Transceiver (NSAT), and  demonstrates proof-of-concept learning applications.
\textcolor{black}{Extreme efficiency in data-driven autonomy hinges on the establishment of (i) energy-efficient computational building blocks and (ii) algorithms that build on these blocks. NSAT is a spiking neural network architecture designed on these assumptions, using neural building blocks that are constructed from algorithmic principles and an event-based architecture that emphasizes locally dense and globally sparse communication~\cite{Park_etal17_hieraddr}.}

To achieve extreme efficiency in dedicated implementations, the NSAT framework consists of neural cores that take advantage of tractable linear neural model dynamics, multiplier--less design, fixed--width representation and event-driven communication, while being able to
simulate a wide range of neural and plasticity dynamics. 
Each NSAT core is composed of state components that can be flexibly coupled to form multi-compartment generalized integrate-and-fire neurons, allowing the implementation of several existing neural models
(Fig. \ref{Fig:nsat_intro}).
The state components forming the neuron can be interpreted as somatic potential, dendritic potential, synaptic currents, neuromodulator concentration or calcium currents, depending on its interactions with other state components or pre-synaptic neurons.
The communication between cores and event-driven sensors is routed via inter-core spike events.

While several neuromorphic VLSI circuits for synaptic learning exist \cite{Qiao_etal15_recoon-l,Pfeil_etal12_4-bisyna,Arthur_Boahen06_learsili,Azghadi_etal15_progspik},
our framework is novel in that it is equipped with a flexible and scalable event--based plasticity rule that is tightly guided by algorithmic considerations and matched to the neuron model. 
Scalability is achieved using only forward lookup access of the synaptic connectivity table  \cite{Pedroni_etal16_forwtabl}, permitting scalable, memory-efficient implementation compared to other implementations requiring reverse table lookups or memory-intensive architectures such as crossbar arrays. 
Flexibility in the learning dynamics is achieved using a reconfigurable event--based learning dynamics compatible with three-factor rules \cite{Urbanczik_Senn14_learby}, consistent with other established plasticity dynamics such as STDP \cite{Bi_Poo98_synamodi,Markram_etal12_spikplas}, membrane-voltage based rules \cite{Clopath_etal10_connrefl}, calcium based dynamics 
\cite{Shouval_etal02_unifmode,Graupner_Brunel12_calcplas}, and reinforcement learning \cite{Florian07_reinlear}.

NSAT is a framework intended to guide the design of an optimized digital architecture, which we outline in the Methods and Results sections.
To set sail towards hardware implementations of the NSAT framework and assist algorithmic co-design efforts, we wrote cNSAT, a multi-thread software simulator of the NSAT framework that is behaviorally accurate with respect to the envisioned optimized digital hardware implementation. 
Using the cNSAT simulator, we show that learning in digital NSAT requires fewer SynOps compared to MACs in equivalent digital hardware, suggesting that a custom hardware implementation of NSAT can be more efficient than mainstream computing technologies by a factor equal to the J/MAC to J/Synop ratio.
Furthermore, to verify the viability of a digital implementation, we validated NSAT on a Field Programmable Gate Array (FPGA).

This article is organized as follows: In the Material \& Methods section we describe the neuron model and
its mathematical equations. 
We present the NSAT architecture and software simulator (publicly available under GPLv3 license). 
In the Results section we show that the neuron model can simulate the Mihalas--Niebur neuron and thus demonstrate a rich repertoire of spike behaviors and neural field models. Then, we demonstrate that the NSAT framework supports a type of gradient back--propagation in deep networks, unsupervised learning in spike-based Restricted Boltzmann Machines (RBMs), and unsupervised learning of sequences using a competitive learning.

\section{Materials and Methods}
In this section we introduce the mathematical description of the NSAT framework
and the details regarding its architecture and the main information processing flow. NSAT software
implementation (cNSAT) details are given in the Appendices.

\subsection{Leaky Integrate-and-Fire Neurons as Dynamical Systems}
\label{sec:si-nsat-neuron}
\textcolor{black}{We start the discussion with the leaky integrate--and--fire neuron (LIF) model, given by the following equations}
\begin{subequations}
\begin{align}
	\label{eq:lif}
    \tau_m \frac{\D}{\D t}  V(t)&= -V(t) + R I(t). \\
    \text{If } V(t) &\geq \theta \hspace{1mm} \text{ then } \hspace{1mm} V(t) = V_r \hspace{1mm} \text{ and } \hspace{1mm} s = 1,
\end{align}
\end{subequations}
where $V(t)$ is the neuron's membrane potential, $\tau_m$ is the membrane time constant, $R$ is the membrane 
resistance and \textcolor{black}{$I(t)$ is the driving current, which can be comprised of external current $I_{\text{ext}}$ and/or synaptic ones $I_{\text{syn}}$.}
When the membrane potential is greater or equal to a threshold value ($\theta$), the neuron fires a spike and the membrane potential value at that time step is set to a reset value $V_r$ (resting potential). 

The dynamical properties of LIF neurons can be extended with synaptic dynamics or other internal currents such as calcium channels, potassium channels and other biophysical variables.
For instance, the concentration of some neurotransmitter or ion, $U(t)$, can be captured by the linear dynamics:
\begin{align}
	\label{eq:lif_com}
    \tau_U \frac{\D }{\D t} U(t) &= -U(t) + \sum_{k} \delta(t-t_k), 
\end{align}
where $U(t)$ is the concentration within the neuron cell reflecting for example calcium concentration, although the biological interpretation is not indispensable for the NSAT framework.
The term $\sum_{k} \delta(t-t_k)$ indicates the pre-synaptic incoming spikes to the current post-synaptic neuron ($\delta$ is the Dirac function\footnote{$\delta(t) = \infty$, if $t = 0$ otherwise $\delta(t) = 0$.
In the discrete version we have $\delta(t) = 1$, if $t = 0$ otherwise $\delta(t) = 0$.}).
If we rewrite the summation term as $S(t)=\sum_{k} \delta(t-t_k)$ then the dynamics become the linear system:
\begin{align}
\begin{bmatrix}
	\dot{V}(t) \\
    \dot{U}(t)
\end{bmatrix}
= 
\begin{bmatrix}
  -\frac{1}{\tau_m}  &  0  \\
   0  & -\frac{1}{\tau_U} 
 \end{bmatrix}
\cdot
\begin{bmatrix}
 	V(t) \\
    U(t)
\end{bmatrix}
+
\begin{bmatrix}
	\frac{RI(t)}{\tau_m} \\
    \frac{S(t)}{\tau_{U}} 
\end{bmatrix}.
\end{align}
A generalization of such linear dynamics to $N$ dimensions can be written in the following vector notation:
\begin{align}
\label{eq:matnsat}
    \frac{\D }{\D t} {\bf x}(t)&= {\bf A}{\bf x}(t) + {\bf Z}(t),
\end{align}
where the temporal evolution of state ${\bf x}(t) = (x_0(t), x_1(t), x_2(t), \cdots, x_N(t))$ is characterized by the solution of Eq.~\eqref{eq:matnsat}.
In the equation above, ${\bf A}$ is the state transition matrix and ${\bf Z}(t)$ the time-varying external inputs or commands to the system. Solutions to linear dynamical systems of Eq.~\eqref{eq:matnsat} are given by:
\begin{align}
	\label{eq:solution}
	{\bf x}(t) &= \exp({\bf A}t) {\bf x}(t_0) + \int_{t_0}^{t} \exp({\bf A}(t - \tau)) {\bf Z}(\tau) d\tau.   
\end{align}
Equation~\eqref{eq:solution} can be computed numerically by using the Putzer algorithm 
\cite{Putzer66_avoijord} for computing the matrix exponential.
\textcolor{black}{Numerical solutions of equation~\eqref{eq:matnsat} can be obtained using several numerical integration methods, such as the Forward Euler which is computationally simple, fast,
less expensive than other Runge-Kutta methods which require more operations, and compatible with stochastic differential equations ~\cite{Kloeden_1991_numerical}. Furthermore, even in the case where equation~\eqref{eq:matnsat} is stiff we can adjust the time-step such that the Forward Euler is stable. For these reasons, Forward Euler is a common choice for digital simulations of neural networks \cite{Davies_etal18_loihneur,Zenke_Gerstner14_limito}}

\subsection{NSAT Neuron and Synapse Model}
In continuous form, the NSAT neuron consists of linear dynamics described in general by equation 
\eqref{eq:matnsat} extended with firing thresholds, resets mechanisms and inputs $\mathbf{Z(t)}$ written
in open form:
\begin{subequations}
\label{eq:nsat_n}
\begin{align}
    \frac{\D {\bf x}(t)}{\D t} &= {\bf A}{\bf x}(t) + \underbrace{({\pmb \Xi}(t)\circ{\bf W}(t))\cdot{\bf s}(t) + {\pmb \eta}(t) + {\bf b}}_{Z(t)}.\\
	\text{If } {\bf x}(t) &\ge
    {\pmb \theta} \hspace{1mm} \text{ then } \hspace{1mm} {\bf x}(t) = {\bf X_{r}}, \\ 		
     \text{If } x_0(t) &\ge \theta_0 \hspace{1mm} \text{ then } \hspace{1mm} s_0(t) = \delta(t).
\end{align}
\end{subequations}
The state components $\mathbf{x} = (x_0, ..., x_k)$ describe the dynamics of a neural compartment or variable such as membrane potential, internal currents, synaptic currents, adaptive thresholds and other biophysical variables, although a biological interpretation is not essential. 
${\bf A}$ is the state-transition square matrix that describes the dynamics of each state component and their couplings.
${\pmb \Xi}$ is a random variable drawn from a Bernoulli distribution and introduces multiplicative stochasticity to the NSAT, which is an important feature for learning \cite{Wan_etal13_reguneur,Hinton_etal12_imprneur} inspired by synaptic failures \cite{Vogelstein_etal02_spiktimi,Neftci_etal16_stocsyna}.
\textcolor{black}{Probabilistic synapses support
Poisson-like variability in the 
of spiking neural networks and provide a mechanism for performing highly robust probabilistic inference under noisy and ambiguous conditions~\cite{Moreno-Bote14_poisspik}. Furthermore, stochasticity at the synaptic level accounts for optimizing the energetic efficiency of  neurons~\cite{Levy_Baxter02_enerneur}.} 

${\bf W}$ is the synaptic strength matrix and defines the connectivity and the strength of each connectivity between neurons.
${\bf s}$ is a vector that takes values in $\{0, 1\}$ and registers whether the neuron has spiked.
The $\circ$ symbol defines an element-wise multiplication (or Hadamard product).
${\pmb \eta}$ is the additive normal noise with zero mean and programmable
variance, \textcolor{black}{allowing for shifting the equilibrium of equation~\eqref{eq:nsat_d} and decorrelating repetitive spiking patterns~\cite{tuckwell:2010}. }
And finally, ${\bf b}$ is a constant value that is added to each state component acting as a constant input (\emph{i.e.} current injection from a neuroscience point of view or bias from a machine learning point of view).
When a component $x_i$ crosses its threshold value ($\theta_i$) then it is subject to reset to some predefined value $X_{r_i}$.
Additionally if the zero state of a neuron ($x_0(t)$) crosses its threshold value then that neuron fires a spike as shown in Eq.~\eqref{eq:nsat_n}, with ($s_0(t) = \delta(t)$) and a new setting of $X_{r_0}$. 
After the neuron has spiked, the membrane potential is clamped during a programmable refractory period, during which it is not permitted to fire.
\begin{figure}[!htpb]
	\centering
    \includegraphics[width=.95\textwidth]{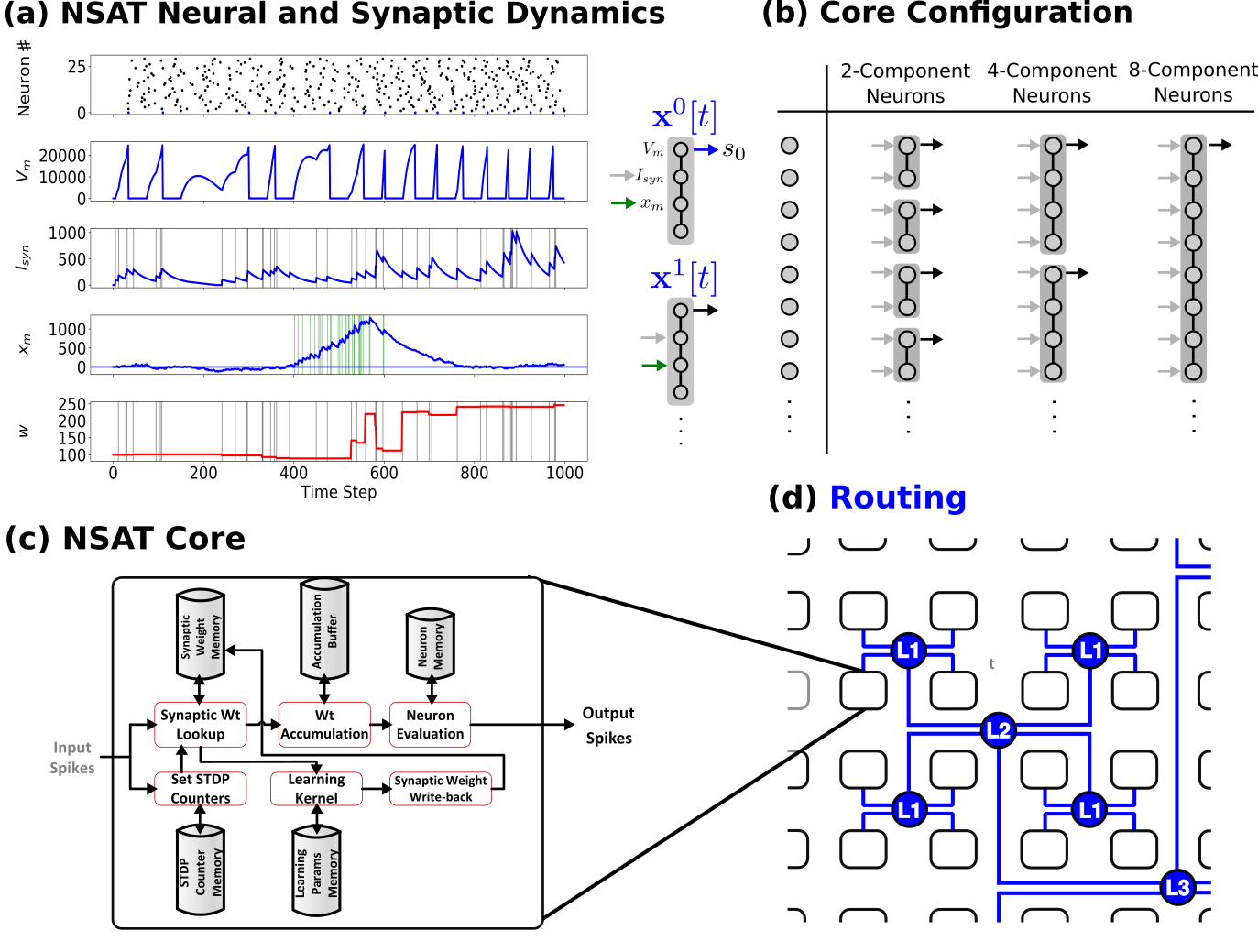}
    \caption{{\bfseries \sffamily The Neural and Synaptic Array Transceiver (NSAT)}.
    {\bfseries \sffamily (a)} Sample run  externally modulated STDP, showing raster plot of 25 neurons and
    detailed temporal dynamics of the four components of the first neuron's (neuron number $0$) state.
    The first component represents the membrane potential ($V_m$), the second component represents the synaptic
    state ($I_{syn}$), while the third is the plasticity modulation ($x_m$). The latter state is driven
    externally by a spike train that is active between time steps 400 and 600, and white noise of constant amplitude.
    {\bfseries \sffamily (b)} NSAT Neurons consist of compartments that can be coupled to trading off the number
    of neurons \emph{vs.} neuron complexity (number of compartments),
    {\bfseries \sffamily (c-d)} NSAT information flow and envisioned layout of the NSAT cores using Hierarchical
    Address--Event Routing (HiAER)~\cite{Park_etal17_hieraddr} for scalable and expandable neural event
    communication with reconfigurable long-range synaptic connectivity.}
    \label{Fig:nsat_intro}
\end{figure}
\subsubsection{Event-driven Synaptic Plasticity and the NSAT Plasticity Model}\label{sec:methods_plasticity}
Spike-Timing Dependent Plasticity (STDP) is a popular learning rule used throughout computational neuroscience models, thanks to empirical evidence and its simplicity.
It is a form of Hebbian learning that modifies the synaptic strengths of connected pre- and post-synaptic neurons based on their spikes firing history in the following way \cite{Bi_Poo98_synamodi,Sjostrom_etal08_dendexci}:
if a post--synaptic neuron generates action potential within a time interval after the pre--synaptic neuron has fired multiple spikes then the synaptic strength between these two neurons potentiates (causal update, long--term potentiation--LTP).
On the other hand if the post-synaptic neuron fires multiple spikes before the pre--synaptic neuron generates action potentials within that time--interval then the synapse depotentiates (acausal update, long--term depression--LTD). 

Like Hebb's rule, STDP is an unsupervised rule that depends on pre-synaptic and post-synaptic factors (here spike times), and so STDP alone is impractical for learning with reward or error signals extrinsic to the STDP neuron pairs. 
On the other hand, three factor rules solve this problem by adding a factor indicative of reward, error, gradients  provided extrinsically or through other neural states \cite{Urbanczik_Senn14_learby,Clopath_etal10_connrefl}. 
Several theoretical work underline that gradient descent on spike train distances or classification loss indeed take the form of such three factor rules \cite{Neftci_etal17_evenranda,Zenke_Ganguli17_supesupe,Pfister_etal06_optispik,Urbanczik_Senn14_learby}.
These results also indicate that optimal gradient descent learning rules involve continuous-time dynamics. 
However, because continuous-time updates are prohibitive in digital hardware, one must resort to event-based learning such as STDP.

To implement multiple learning scenarios in a fully event-based fashion with minimal memory overhead, NSAT follows a modulated, index-based STDP rule.
Index-based architectures are memory-efficient with realistic and practical sparse connectivities, but are challenging to implement in neuromorphic hardware because synaptic memory is typically localized at the pre-synaptic neurons, and so causal updates require reverse look-up tables or reverse search for the forward table at every spike-event.
Although reverse lookups are not an issue in crossbar memories (they are compatible with the data structure associated with the crossbar), they can incur a significant memory overhead for non-dense connectivities and are not considered here.

Recent implementations of STDP in the Spinnaker hardware use dedicated synaptic plasticity cores \cite{Galluppi_etal15_framplas} for implementing STDP. 
While this approach gives additional flexibility in the STDP learning rule, it relies on relatively large SDRAMs and more communication for its realization. 
Furthermore, it has the disadvantage of segregating synaptic memory from neural states, which as argued above, may contain important information for learning.

To mitigate these problems, NSAT uses a forward table--based, pre--synaptic event-triggered, nearest--neighbor STDP rule \cite{Pedroni_etal16_forwtabl} coupled with the neuron dynamics.
This method implements both causal and acausal weight updates using only forward lookup access of the synaptic connectivity table. A single timer variable for each neuron is sufficient to implement this rule, permitting implementation that requires only $\mathcal{O}(N)$ memory, where $N$ is the number of neurons. 
\textcolor{black}{The basic nearest-neighbor STDP~\cite{Sjostrom_etal08_dendexci} is recovered in the case of refractory periods greater than the STDP time window, and otherwise it closely approximates exact STDP cumulative weight updates.}

This method is related to the deferred event-driven (DED) rule used in Spinnaker \cite{Jin_etal10_implspik}, which does not allow the pre-synaptic spike to trigger the STDP until a predetermined time limit is reached. 
The time that a pre-synaptic spike occurred is recorded as a time-stamp and is used in the future once the missing information from the future spikes has been made available (post-synaptic neurons have fired action potentials). 
Such STDP schemes are called ``pre-sensitive'', as STDP takes place only when a pre-synaptic neuron fires an action potential. 

Similar to DED, the three-factor NSAT STDP learning rule implements a pre-sensitive scheme: 
The NSAT framework learning rule keeps track of the spike times using a time counter
per neuron.
When a pre-synaptic neuron fires then all the corresponding acausal STDP updates are triggered and the post-synaptic weights are updated based on a linear or exponential approximation STDP kernel (see Fig.~\ref{Fig:stdp_kernel} in
Appendix~\ref{sec:soft_implem}).
If the counter of the pre-synaptic neuron expires, then only the causal STDP update takes place.
As long as the counter has not expired and post-synaptic neurons fire within the STDP time-window then the acausal
updates are computed. If, now, a new spike from the pre-synaptic neuron is emitted then causal updates are 
computed. 

To enable learning using extrinsic and intrinsic modulation, the NSAT three-factor learning rule is modulated by the neural state components, which enables modulation based on continuous dynamics using both local and global information.
Since NSAT connectivity allows extrinsic inputs to drive its state components, modulation can be driven by extrinsic rewards, error or the dynamics of entire neural population.  
The mathematical formulation of the NSAT three-factor STDP learning rule is given by:
\begin{subequations}
\label{eq:learning}
\begin{align}
	\epsilon_{ij}(t) &= x_m^j(t) \Big( K(t - t_i) + K(t_j - t) \Big), \\
  	\frac{\D }{\D t} {\bf w}_{ij}(t) &= \epsilon_{ij}(t) \delta_j(t),
\end{align}
\end{subequations}
where $\epsilon_{ij}(t)$ is the eligibility of the synapse between the $i$--th pre--synaptic neuron and the $j$--th post-synaptic neuron update, $K(\cdot)$ is the STDP learning window, $t_i$ is the last time that the $i$--th pre--synaptic neuron fired a spike, and $t_{j}$ is the last time that the post-synaptic neuron fired a spike.
The kernel (or learning window) $K(t-t_i)$ refers to the causal (positive) STDP update and the term $K(t_j -t)$ refers to the acausal (negative) STDP update. 
$x_m^j(t)$ is the $m$--th state component of the post--synaptic neuron that dynamically modulates the amplitude of the STDP kernel. 

\textcolor{black}{
A remark with respect to learning rule~\eqref{eq:learning} is that rate-based learning schemes can be implemented on NSAT, as NSAT neurons are compatible with firing rate neurons.
To achieve this, NSAT neurons configured as integrators to read-out estimates of the firing rate, and the synaptic plasticity rule can be configured as a membrane voltage-modulated learning rule.
Another example of rate-based learning is the BCM learning 
rule~\cite{Bienenstock_etal82_theodeve}, from which one can derive a modulated STDP rule compatible with NSAT assuming stochastic firing of the pre- and post-synaptic neurons \cite{Izhikevich_2003}.
The latter can be realized using additive or multiplicative noise. }

\subsection{Difference Equations of NSAT (Quantized) Framework}
The NSAT software simulator consists of discrete-time versions of the above equations, based on fixed point arithmetics without any multiplications. 
The continuous--time dynamics of NSAT described by Eq.~\eqref{eq:nsat_n} and Eq.~\eqref{eq:learning} are rewritten here in a discrete (quantized) form:
\begin{subequations}
\label{eq:nsat_d}
\begin{align}
	  {\bf x}[t+1] &= {\bf x}[t] + {\bf A} \diamonddiamond {\bf x}[t] + 
      			   + ({\pmb \Xi}[t]\circ{\bf W}[t])\cdot{\bf s}[t] 
        		   + {\pmb \eta}[t] + {\bf b}. \\
       \text{If } {\bf x}[t+1] &\ge {\pmb \theta} \hspace{1mm} \text{ then } \hspace{1mm}
   			{\bf x}[t+1] \leftarrow {\bf X_{r}}. \\ 		
      \text{If } x_0[t+1] &\ge \theta_0 \hspace{1mm} \text{ then } \hspace{1mm} s_0[t+1] \leftarrow 1, 		     
\end{align}
\end{subequations}
where the entries of matrices ${\bf A}$ and ${\bf b}$ are integer constants, and ${\pmb \eta}$ is the variance of the additive noise.
More details regarding the parameters are provided in Appendix~\ref{sec:soft_implem} and in the SI.

The binary operator $\mathcal{D}(\cdot, \cdot):GF(2^n) \rightarrow GF(2^n)$ (or $\diamonddiamond$), where 
$n=4$ or $n=5$, is defined as
\begin{align}
	\label{eq:diamond}
    \mathcal{D}(a, x)= a \diamonddiamond x =  \left\{
\begin{array}{ll}
       sign(-a\diamond x)& \text{if }  d(a,x) \neq 0  \text{ and } a = 0 \\
       a & \text{otherwise},
\end{array} 
\right. 
\end{align}
(described also by Algorithm~\ref{algo:round} in Appendix~\ref{sec:algorithmic}) plays the role of 
a multiplication implemented with bit shift operations. In particular, it ensures that all state
components leak towards the resting state in the absence of external input (current).

The binary operator $d(\cdot, \cdot):GF(2^n) \rightarrow GF(2^n)$ (or $\diamond$), where $n=4$ or $n=5$, defines a
custom bit shift. It performs a multiplication by power of two using only bitwise operations, and it is defined as
\begin{align}
	\label{eq:small_diamond}
        d(a, x)= a \diamond x = \left\{
\begin{array}{ll}
       x << a & \text{if }  a \geq 0 \\
       sign(x)(|x| >> -a) & \text{otherwise}
\end{array} 
\right. 
\end{align}
(see also Algorithm~\ref{algo:diamond} in Appendix~\ref{sec:algorithmic}).

The reason for using $\diamond$ rather than left and right bit shifting is because integers stored using a two's complement representation have the property that right shifting by $a$ values such that $x>-2^{a'}, \forall a'<a$ is $-1$, whereas $0$ is expected in the case of a multiplication by $2^{-a}$. The $\diamond$ operator corrects this problem by modifying the bit shift operation such that $-2^{a'}\diamond{a}=0,\,\forall a'<a$. 
In addition, such multiplications by powers of $2$ have the advantage that the parameters are stored on a logarithmic scale, such that fewer bits are required to store parameters. 
For example, $- (-3\diamonddiamond x_0)$ is the NSAT equivalent of $-2^{-3} x_0[t]$.
A logarithmic scale for the parameters is suitable since solutions to the equations consist of sums of exponentials of these parameters (\refeq{eq:solution}).

The learning rule given by Eq.~\eqref{eq:learning} is also discretized:
\begin{subequations}
\label{eq:learning_d}
\begin{align}
   \epsilon_{ij}[t] &= x_j^m[t] \diamond \Big( K[t - t_i] + K[t_j - t] \Big), \\
          w_{ij}[t+1] &= \text{Clip}(w_{ij}[t] + \underbrace{\epsilon_{ij}[t] s_j[t]}_{\Delta w_{ij}}).
\end{align}
\end{subequations}
Where $\text{Clip}(x) = \max\{w_{\text{min}}, \min\{x, w_{\text{max}} \}\}$ clips its first argument to within
the range $[w_{\min},w_{\max}]$ dictated by the fixed point representation of the synaptic weights at every time step. 

In addition, the weight updates can be randomized using a discretized version of randomized rounding \cite{Muller_Indiveri15_rounmeth}, which interprets the $r$ least significant bits of $\Delta w$ as a probability, as follows:
\begin{equation}
\label{eq:learning_dR}
  \Delta w^r_{ij} = (\Delta w \gg r) + \begin{dcases*}
          1 & \text{if } random(0,1) \text{$<$} p \\
          0 & \text{otherwise}.
      \end{dcases*},
\end{equation}
where $p$ is the number formed by the $r$ least significant bits of $\Delta w_{ij}$.

Figure \ref{Fig:nsat_intro}(a) shows an example of the NSAT learning rule (Eq.~\eqref{eq:learning_d}).
In this example, each neuron consisted of 4 components. The first and second component correspond to classical leaky 
integrate and fire dynamics with current-based synapses. The third state component driven externally to modulate the STDP 
update. As a result most weights updates concentrate around high modulation states.

\subsection{The NSAT Architecture}

\begin{figure}[!tbph]
	\centering
    \includegraphics[width=.48\textwidth]{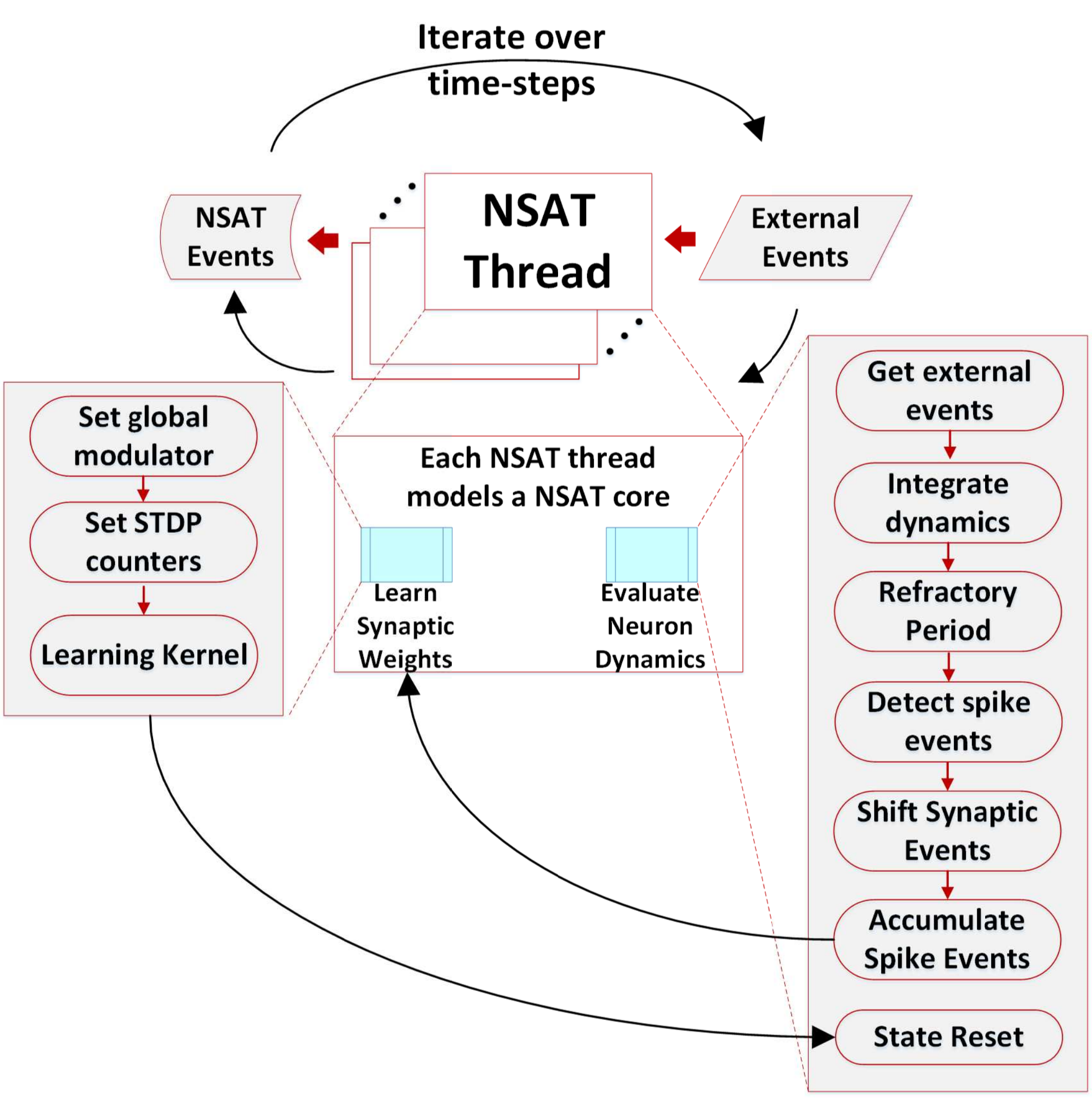}
    \caption{{\bfseries \sffamily The NSAT Architecture.} Multiple NSAT threads receive external events and 
    generate event responses in two stages. Each thread consists 
    of two major instruction sets (cyan boxes), one for
    evaluating neuron dynamics and the other for implementing an event-based STDP learning rule. The instructions
    are shown in the two large gray boxes. Arrows indicate the flow of the information within an NSAT thread during 
    simulation. }
    \label{Fig:nsat_arch}
\end{figure}

Figures~\ref{Fig:nsat_intro}(c), (d), and~\ref{Fig:nsat_arch} illustrate the NSAT architecture consisting of multiple interconnected cores (or threads each simulating one NSAT core). Only addresses of a neuron's spike are transmitted in inter-- and intra--thread communication.
At every simulation time step (or tick) each thread runs independently, executing NSAT dynamics in two stages.
In the first stage each thread integrates the neural dynamics of its neurons based on Eq.~\eqref{eq:nsat_d} without accumulating the synaptic inputs on the neuron state. At that stage all the threads are synchronized (thread barrier)
and then they detect new spike-events and transmit them accordingly to their destinations.
All the detected inter- and intra-core spike events at time $t$ are made available to the next time step ($t+1$).
After the distribution of all the spikes (intra- and inter-core) all the threads are synchronized once again 
before proceed to the next stage. 

In the second stage, the detected spike events (including the external ones) are accumulated onto the neural states components $x_i$ according to $({\pmb \Xi}[t]\circ{\bf W}[t])\cdot{\bf s}[t]$. 
Synaptic weights are multiplied with a predefined constant (implemented as a bit shift operation) to trade off precision and range limitations imposed by fixed point integer arithmetic. 
Our previous work~\cite{Neftci_etal17_evenrand} and Drop Connect~\cite{Wan_etal13_reguneur} showed that a probability of $\frac{1}{2}$ work best as multiplication constant. Consequently, we use a blank-out factor as close as possible to $\frac{1}{2}$ throughout our simulations. 
The blank-out factor does not directly affect
the required weight precision.
When the learning is enabled, Eq.~\eqref{eq:learning_d} is computed.
First, threads compute the causal and then the acausal part of the STDP learning curve.
After learning, the STDP counters of neurons that have spiked are set to their new values (either the last time that a neuron spiked, or a neuron clock starts ticking until expiration).
The final steps in the second stage perform  update of modulator dynamics ($x^m_i$) and  reset of the neuron state components that spiked.
The modulator state component ($x_m$) adjusts the amplitude of the STDP function as described in Eq.~\ref{eq:learning_d}.
Algorithm~\ref{algo:nsat_sof} provides in pseudo-code the flow of the NSAT operations.
Furthermore, in the Appendix~\ref{sec:soft_implem} we provide more details regarding the data structures,
simulation details and in the SI the parameters for all of our results presented in the next section. 
\begin{algorithm}[!tbph]
	\begin{algorithmic}
    	\Require Synaptic Weights, Parameters, Learning Parameters
        \Ensure Spike events, States, Synaptic Weights
        \ForAll{$p$ in $\{\text{Threads}\}$}
        	\For{$t \gets 1 \ldots t_{\text{final}}$}
        		\ForAll{$i$ in $\{\text{Neurons}\}$} 
        		\State spike\_list $\gets$ external\_events
            	\State ${\bf x}^i[t] \gets {\bf x}^i[t] + {\bf A} \diamonddiamond {\bf x}^i[t] + {\pmb \eta}[t] + {\bf b}$
            	\If{$ \text{ref\_period}^i > 0$}
            		\State $\text{ref\_period}^i \gets \text{ref\_period}^i - 1$
            		\State ${\bf x}^i[t] \gets {\bf X}^i_{\text{reset}}$
            	\EndIf
            	\If{Spike is Enabled}
                	\If{${\bf x}^i[t] \geq {\pmb \theta^i}$}
                    	\State spike\_list $\rightarrow$ id $\gets i$
                		\State spike\_list $\rightarrow$ ts  $\gets t$
                    	\State $s^i[t] = 1$\;
                    \EndIf
                \ElsIf{Adaptive $\theta$ is Enabled}
                	\If{$x^i_0[t] \geq x^i_1[t]$}
                    	\State spike\_list $\rightarrow$ id $\gets i$
                		\State spike\_list $\rightarrow$ ts  $\gets t$
                    	\State $s^i[t] = 1$
                    \EndIf
            	\EndIf
            	\State ${\bf x}^i[t] \gets {\bf x}^i[t] ({\pmb \Xi}[t]\circ{\bf W}[t])\cdot{\bf s}^i[t]$
            	\State ${\bf W}[t] \gets {\bf W}[t] \circ G^i$
            	\If{Learning is Enabled}
            		\State Compute equations~\eqref{eq:learning_d} and \eqref{eq:learning_dR}
           		\EndIf
				\If{$s^i[t] == 1$ and Reset is Enabled}
            		\State ${\bf x}^i[t] \gets {\bf X}^i_{\text{reset}}$
                    \State $\text{ref\_period}^i \gets {\bf X}^i_{\text{ref} }$
            	\EndIf
            	\EndFor
           \EndFor
        \EndFor
	\end{algorithmic}
\caption{Algorithmic (software) NSAT implementation (see text for more details).}
\label{algo:nsat_sof}
\end{algorithm}

\subsection{NSAT Hardware Architecture}

A synchronous digital architecture with the same functionality as the cNSAT was written in Verilog and its functionality was validated by emulating the same on FPGA.
This section provides an overview of the architecture and provides an idea on the potential power savings that result from optimized NSAT data-structure and functions. 
\begin{figure}
	\centering
    \includegraphics[width=0.65\textwidth]{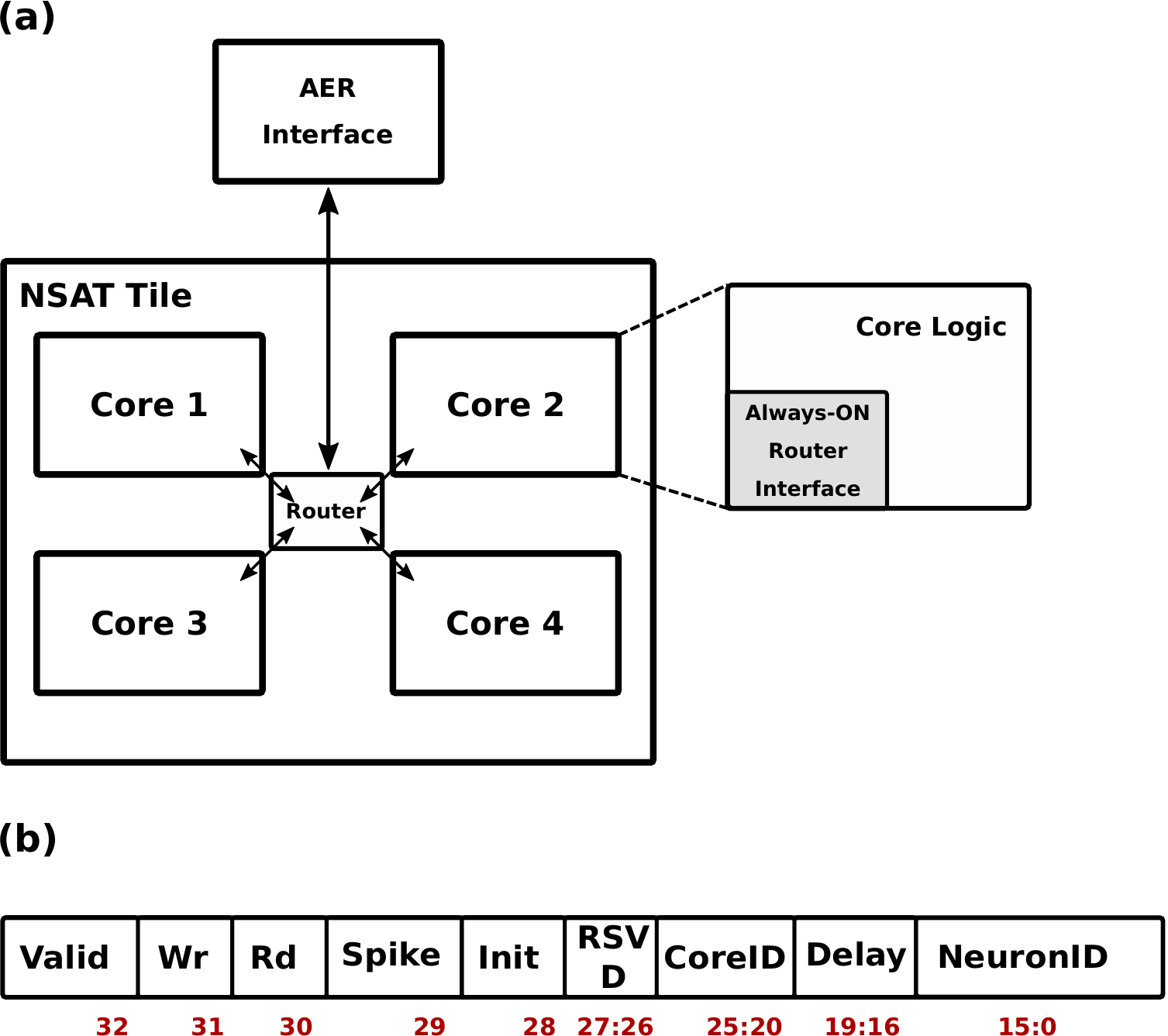}
    \caption{{\bfseries \sffamily Top-level NSAT Architecture.} The NSAT tile 
    block ({\bfseries \sffamily a}) and the organization of a spike packet 
    ({\bfseries \sffamily b}).}
    \label{Fig:top-level-nsat}
\end{figure}

Figure~\ref{Fig:top-level-nsat}(a) shows the top level organization
of the NSAT architecture. We refer to this as a single NSAT tile. Note that such tiled 
architecture has been proposed earlier in the
context of neuromorphic hardware with multi-tile communication enabled through a
hierarchical AER communication fabric~\cite{Park_etal17_hieraddr}. The contribution 
of this work therefore
focused on specifics of the digital implementation inside each tile. Each tile is 
hierarchically organized into four NSAT cores, which communicate via a packet-switched
router conforming to the Address Event Representation (AER) protocol \cite{Lazzaro_etal93_siliaudi}.
The AER packets are routed from/to the primary AER interface at each tile to each core following
a wormhole routing strategy implemented in the router.
The digital implementation of the router is inspired from~\cite{Vangal_et_al:2007}, 
which is adopted to work on single-flit packets.
These packets have the format as shown in Fig.~\ref{Fig:top-level-nsat}(b). 

As shown in Fig.~\ref{Fig:top-level-nsat}(b),
the packet is functionally diverse. It can act as (i) a memory write
packet – to initialize the weight memory and the configuration register inside each core.
The Neuron and Delay fields then carry the address and the payload for writing to the memory
location, (ii) a memory read packet – to read from a memory location in a given core.
The payload is then the memory address. In response to a memory read packet, the core
responds by sending out the data being read from the memory location.
(iii) A spike packet, representing a spike from another core or tile, carries the destination core and neuron addresses along with an axonal delay information. This conforms to the
general AER definition where an event is tagged with a destination address.
Each core is logically divided into an always-ON router interface and the core logic
which is active only at the arrival of an input spike or at the beginning of each
time-stamp. If no spike is present in a given time-stamp, only the neuron dynamics are
evaluated and the core logic is thereafter put in a low-power retention mode. 

\subsubsection{NSAT Core Architecture}

\begin{figure}[htpb!]
	\centering
    \includegraphics[width=1.\textwidth]{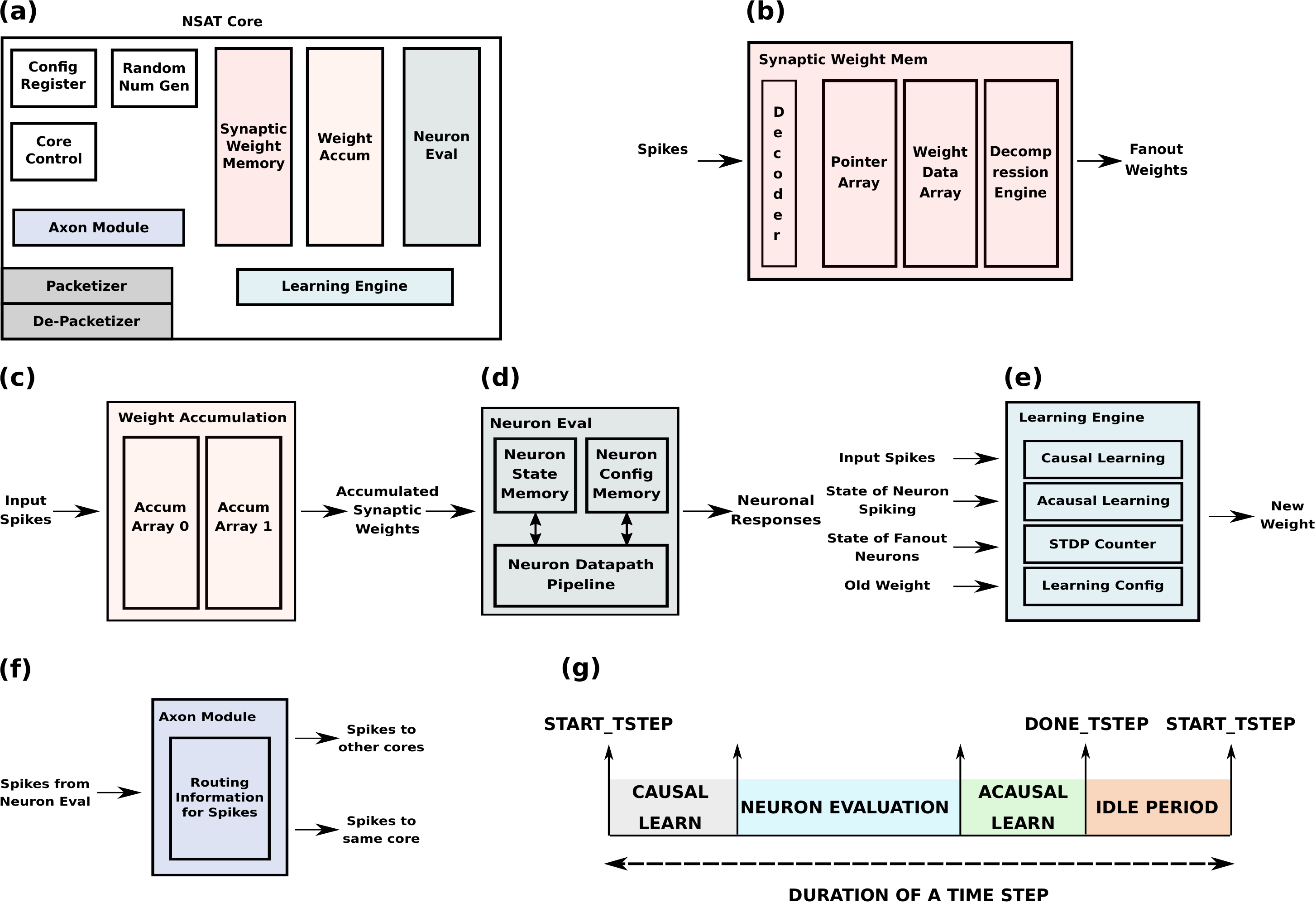}
    \caption{{\bfseries \sffamily NSAT Hardware Core Architecture.} 
    ({\bfseries \sffamily a}) The architectural organization of an NSAT hardware core 
    and the different modules constitute the core. 
    ({\bfseries \sffamily b}) Synaptic weight memory module, 
    ({\bfseries \sffamily c}) synaptic weights accumulation module, 
    ({\bfseries \sffamily d}) neuron evaluation module,
    ({\bfseries \sffamily e}) learning engine, and 
    ({\bfseries \sffamily f}) neuron axon module,
    ({\bfseries \sffamily g}) time-step duration indicating the sequence of operations
    in each core.}
    \label{Fig:hard_arch_fig}
\end{figure}

Figure~\ref{Fig:hard_arch_fig} shows the detailed breakdown of each core.
The Always-ON (AON) router interface consists of two channels - packetizer and de-packetizer
corresponding to the outgoing and incoming streams of packets. The AON module also generates
a gated clock for the rest of the NSAT core. In absence of input activity, the core clock is
gated to prevent dynamic power dissipation. Each NSAT core contains logic and memory to map
$512$ $8$-component state neurons which can also be reconfigured as $4096$ $1$-component 
state neuron. Following are the primary components of the NSAT core.

The synaptic weight memory is physically the largest of the NSAT core modules, with $128\, \mathrm{KB}$ of synaptic weight storage.
Input to this module are spikes and output are its fanout weights.
In addition to the synaptic storage, decoder logic selects the appropriate memory array for weight retrieval.
In order to store weights for sparse fanouts, we have implemented a compressed storage scheme for the weights. 
The weight memory is divided into a pointer array which stores the pointers to the weight data array, while the weight data array stores the actual weights.
The compression scheme we used is Run Length Encoding (RLE) which skips zero weights corresponding to missing connections.
The pointer memory is also useful for storing pointers for weight parameters that are shared across multiple neurons (\emph{e.g.} in convolutional filters).
A decompression engine following the weight data array decompresses the weights before sending them out for accumulation to destination neurons. 
The organization is illustrated in Fig.~\ref{Fig:hard_arch_fig}(b).

The fanout weights obtained are accumulated on the destination neurons in the weight accumulation module. 
While, one memory array is used to store and update the partial sums from weight accumulation in the current time-stamp, another array is used to feed the accumulated weights from the previous time-stamp to the neuron evaluation unit. 
This approach decouples weight look-up and accumulation from neuron evaluation. 
The total memory size is proportional to the number of neurons, number of state components/neuron and the number of bits per component. 
Figure~\ref{Fig:hard_arch_fig}(c) illustrates the organization of the weight accumulation module.  

The neuron evaluation block  evaluates the neuron dynamics in the NSAT core.
It receives the accumulated weights from the weight accumulation block.
The previous state components are stored in the neuron state memory in the neuron evaluation 
block.
The neuron configuration parameters are also locally stored in the neuron evaluation block. 
The NSAT neural dynamics is implemented as a $4$-stage pipeline where all the state components of a neuron are evaluated in parallel and all the neurons are evaluated in a time-multiplexed fashion. 
The output from the block are neuronal responses, \emph{i.e.} spikes. Figure~\ref{Fig:hard_arch_fig} (d) shows the block diagram of the neuron evaluation unit. 

The learning engine implements multiple learning algorithms as available in cNSAT (Fig.~\ref{Fig:hard_arch_fig}(e)).
This includes weight update corresponding to conventional STDP as well as the state dependent weight update.
For STDP weight update, both the causal and acausal pipelines were used, using a forward table-based pre-synaptic event-triggered STDP (as
described in Sec. \ref{sec:methods_plasticity}), accordingly to which the state dependent weight updates use only the acausal pipeline.
A dedicated latch-based memory module implements the STDP counters.
The latch-based design instead of a register file/SRAM based approach allows multiple counters to be updated at the same time corresponding
to multiple spikes. 
In addition to the input spikes (from internal and external to the core), in order to perform state-dependent weight update, the learning engine reads (from the neuron evaluation unit) the state of its fanout neurons.
The learning configuration parameters are stored in a small memory in the learning engine. 

The axon module holds the routing information to route spikes into the same core or to other cores (Fig.~\ref{Fig:hard_arch_fig}(f)).
The routing information is stored per each neuron that is mapped to a given core.
A flag in the routing table indicates whether a spike generated in the core is routed back or routed outside. 
If it is expected to be routed back, it is inserted into the queue at the input of the synaptic weight memory module.
It then performs weight look-up and accumulation in a manner similar to other spikes coming from other cores.
For spikes destined for other cores, the output from the axon module is routed to the packetizer unit in the AON router interface block.

Stochastic learning (stochastic synapses and randomized rounding) have proven to be extremely effective in the NSAT framework (see Supervised and Unsupervised sections in Results).
To support randomness we implement a robust Linear-feedback Shift Register (LFSR)-based pseudo-random number generator \cite{Tkacik02_hardrand}.
There are four individual uniform random number generators, which are combined to generate a normal random sequence which can also be used as a random noise on the neuron membrane potential~\cite{Neftci_etal17_evenranda}. 

The control unit oversees the overall control of operations in the NSAT core.
It is implemented as a state-machine which is responsible for triggering smaller state-machines in each individual block in the design.
The sequence of operations in each core follow the behavior as shown in Fig.~\ref{Fig:hard_arch_fig}(g).
The beginning of a time step is indicated by the \texttt{start\_tstep} signal, which is a global signal that is broadcasted from the main control center (PC in this case), while the $4$ cores in the NSAT tile act as slave accelerator cores.
Each core indicates the completion of its neuron evaluation and learning periods by sending out the \texttt{done\_tstep} signal to the control center. 
Once \texttt{done\_tstep} is received from all cores, the control center waits for any global time-step constraint (e.g. simulated time-constant of 1ms) to elapse before sending out the \texttt{start\_tstep} for the next time-step.
This simple approach allows us to achieve multi-core synchronization which can be easily scaled to multiple NSAT tiles. 

In addition, the control unit also stores spikes which are attributed with non-zero delays.
An array with a size corresponding to the total neuron space $\times$ max future delay time-steps in the control unit stores spikes to be retrieved and used at a future time-step. 
Configuration parameters corresponding to overall mapping of neurons and learning strategy (STDP or state-based) are stored in these global configuration registers.

\subsubsection{Validation of the Architecture}


\textcolor{black}{
FPGA was used as a means of validating the NSAT architecture and pre-Si demonstration of NSAT software. 
It is also intended to capture performance statistics of various parts of the NSAT pipeline (e.g. NSAT
dynamics, learning module) which would otherwise be extremely slow to capture using RTL simulations.
We intend to leverage these statistics for designing better power-management scheme for the NSAT ASIC.
The NSAT mapping on the FPGA consumes all types of resources including logic, DSP and memory. DSP
utilization is low ($<1\%$) since NSAT does not use any large multiplier, but only accumulators and 
shifters. Memory utilization is high ($80$ BRAMs), primarily due to large synaptic weight memory, pointer
memory, neuron state table. 
One quarter of the logic resources is used to map a single NSAT tile. We expect logic resources and 
routing (interconnect resources) to be the limiter to mapping multiple tiles. The design was synthesized
for a clock frequency of $200\, \mathrm{Mhz}$, which was found enough for a real-time demonstration of
spiking neural networks (SNN) based inference workload. At this target frequency, all timing paths
were satisfied. However the most critical path was found in the logic for neuron dynamics for a 
$8$-state neuron scenario. Since the goal of the FPGA mapping was only emulation, no power measurement
was done. }

\begin{figure}[htpb!]
	\centering
    \includegraphics[width=0.5\textwidth]{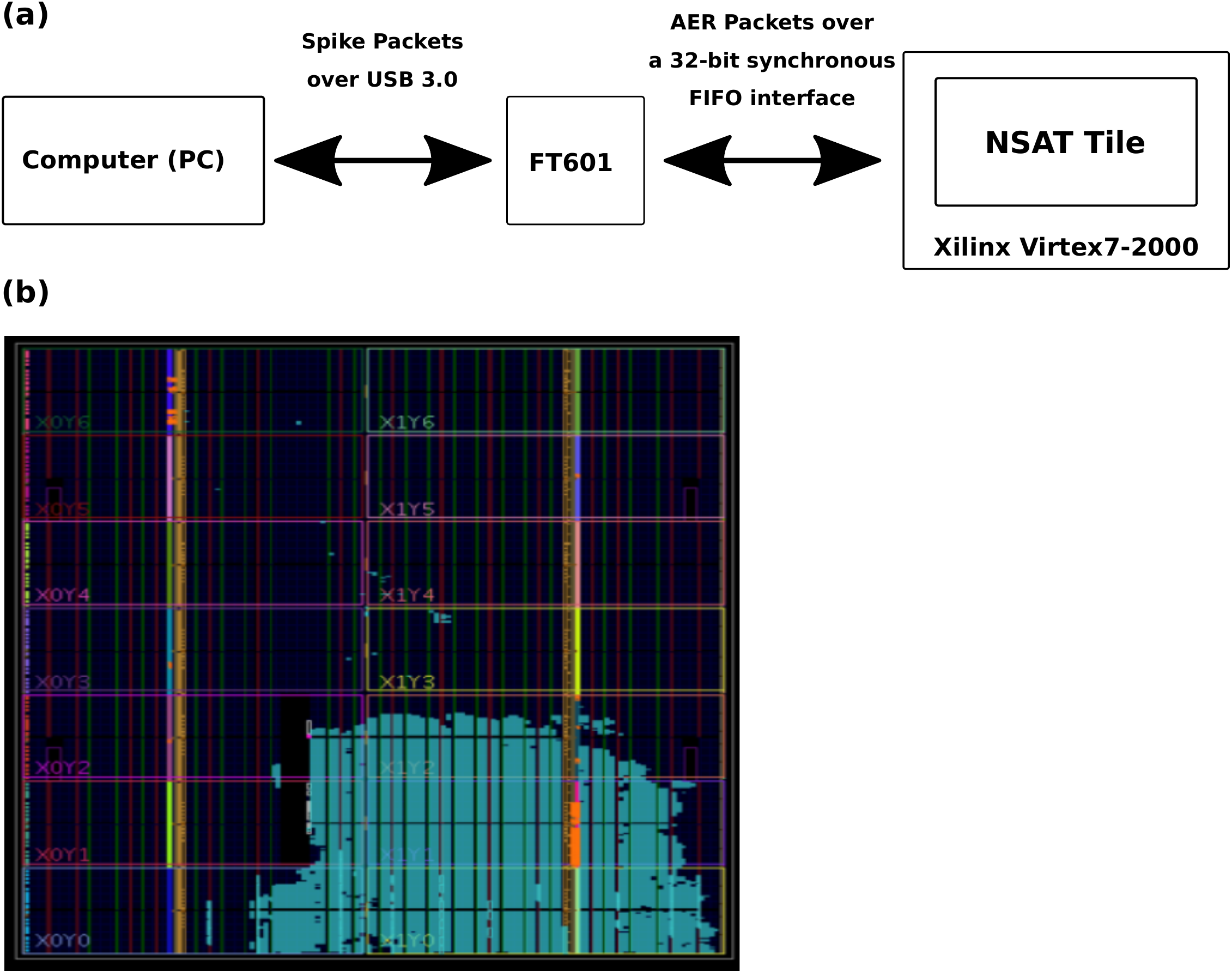}
    \caption{{\bfseries \sffamily FPGA-based emulation of NSAT.}
    ({\bfseries \sffamily a}) Flow of information from a computer to the Xilinx FPGA board.
    ({\bfseries \sffamily b}) NSAT tile mapped to a Virtex-$7$ $2000$T device. }
    \label{Fig:xilinx_nsat}
\end{figure}

\section{Results}

In this section we demonstrate the NSAT capabilities by performing five different tasks using the cNSAT simulator.
First, we show that NSAT supports a wide variety of neural responses, such as tonic, bursting and phasing spiking. The
second task is a simulation of Amari's neural fields \cite{Amari77_dynapatt} in three different applications: stationary
``bump'' solutions, target selection and target tracking.
Then, we illustrate three learning tasks in supervised and unsupervised settings. 

\subsection{NSAT Neuron Dynamics Support a Wide Variety of Neural Responses}
The Mihalas--Niebur neuron (MNN) model \cite{Mihalas_Niebur09_geneline} is a linear leaky integrate-and-fire neuron that is able to capture a wide spectrum of neural responses, such as tonic spiking, bursts of spikes, type I and type II spike responses.
Here, we show that the NSAT neuron model can implement the MNN model and thus simulate a similar spectrum of neural responses.

The MNN model consists of $N+2$ equations, where the first two equations describe the membrane potential and an adaptive threshold, respectively.
The remaining $N$ equations define internal currents of the neuron.  
The subthreshold dynamics of MNN neuron are given by, 
\begin{subequations}
\label{eq:mnn}
\begin{align}
	\frac{\D V(t)}{\D t} &= \frac{1}{\mathrm{C_m}} \Big(I_e - g(V(t) - E_L) + \sum_{j=1}^{N}I_j(t)\Big), \\
    \frac{\D \Theta(t)}{\D t} &= a(V(t) - E_L) - b(\Theta(t) - \Theta_{\infty}), \\
    \frac{\D I_j(t)}{\D t} &= -k_j I_j(t), \qquad j = 1, \ldots, N,
\end{align}
\end{subequations}
where $V(t)$ is the membrane potential of the neuron, $\Theta(t)$ is the instantaneous threshold, $I_j(t)$ is the $j$-th internal current of the neuron. 
$\mathrm{C_m}$ is the membrane capacitance, $I_e$ is the external current applied on the neuron, $g$ is a conductance constant, $E_L$ is a reversal potential. 
$a$ and $b$ are some constants, $\Theta_{\infty}$ is the reversal threshold and $k_j$ is the conductance constant of the $j$--th internal current.
The MNN neuron generates spikes when $V(t) \geq \Theta(t)$ and updates neural state as 
follows: 
\begin{subequations}
\label{eq:mnn_reset}
\begin{align}
	I_j(t) &\leftarrow R_j \times I_j(t) + P_j, \\
    V(t) &\leftarrow V_r, \\
    \Theta(t) &\leftarrow max\{\Theta_r, \Theta(t) \},
\end{align}
\end{subequations}
where $R_j$ and $P_j$ are freely chosen constants, $V_r$ and $\Theta_r$ are the reset values for the membrane potential and the adaptive threshold, respectively. 

We implement the MNN model using the NSAT framework and following the configuration provided in \cite{Mihalas_Niebur09_geneline}.
Therefore, we assume $N=2$ (the number of the internal currents), which has been demonstrated to be sufficient for a wide variety of dynamics~\cite{Mihalas_Niebur09_geneline}.

\begin{figure}[tbph!]
	\centering
    \includegraphics[width=0.5\textwidth]{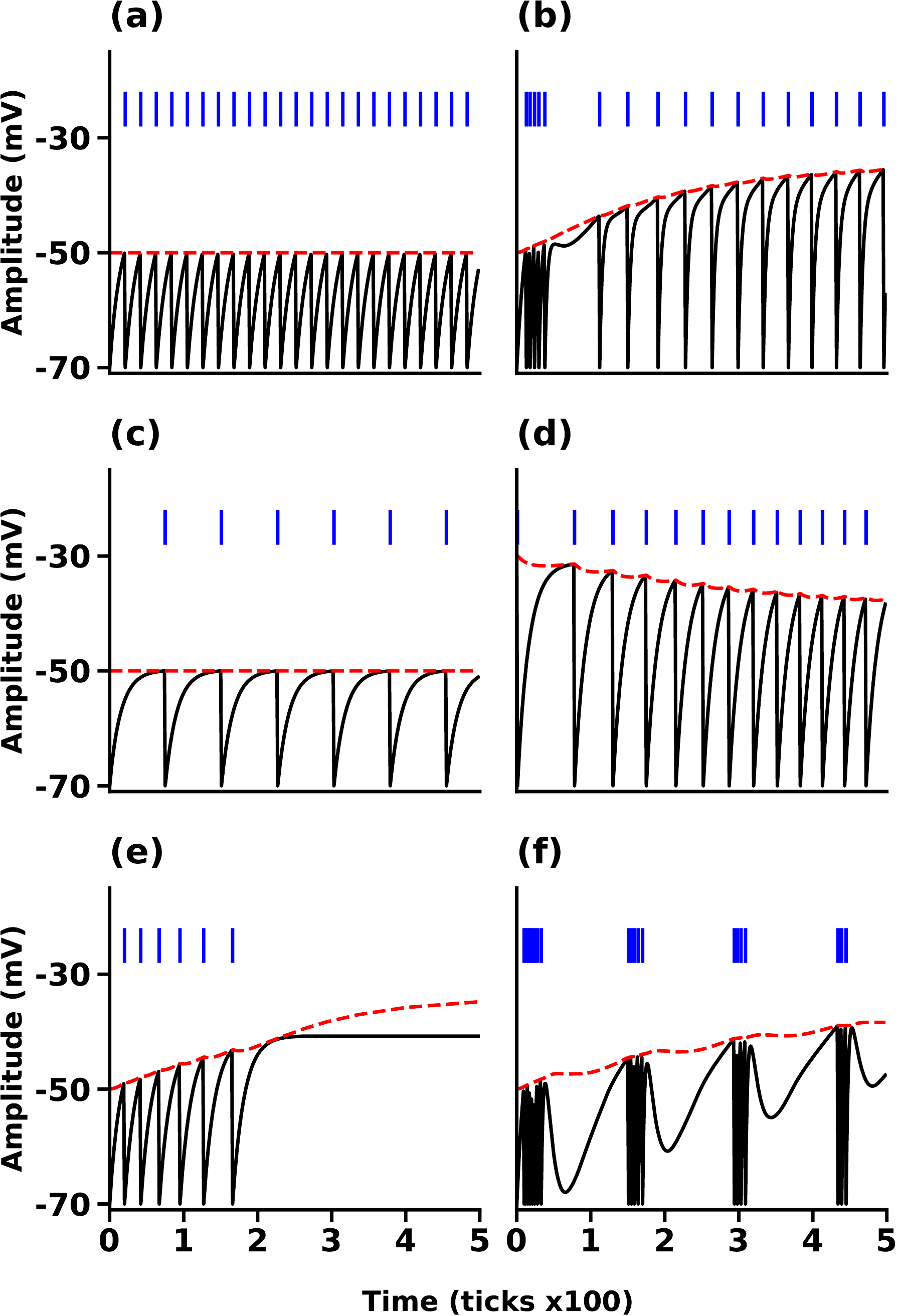}
    \caption{{\bfseries \sffamily NSAT Mihalas-Niebur Simulation.} Results from an NSAT simulation of MNN model. 
    ({\sffamily \bfseries a}) Tonic spiking,
    ({\sffamily \bfseries b}) mixed mode,
    ({\sffamily \bfseries c}) class I neuron,
    ({\sffamily \bfseries d}) class II neuron,
    ({\sffamily \bfseries e}) phasic spiking, and 
    ({\sffamily \bfseries f}) tonic burst.
    Black and red lines indicate the membrane potential (state $x_0(t)$) and the adaptive threshold (state $x_1(t)$),
    respectively. Blue vertical line segments represent spike events.}
    \label{Fig:mnn}
\end{figure}

We simulated the MNN in six different cases, tonic spiking, phasic spiking, mixed mode, class I and II, and tonic bursting.
These six neural responses are important because (i) they are the most frequently used neural responses in the field of computational neuroscience and (ii) in \cite{Mihalas_Niebur09_geneline} all the $20$ different neural behaviors reduce to three different classes in terms of implementation.
Our results produce very similar responses compared to the original MNN ones. 
Figure~\ref{Fig:mnn} illustrates the results of all these six simulations.
The black lines show the membrane potential of the neuron ($V(t)$), the red dashed lines indicate the adaptive threshold ($\Theta(t)$), and the vertical blue line segments show the spike trains for each simulation. 
Some of the simpler neural responses and behaviors provided by Mihalas and Niebur can be achieved by NSAT in a simpler way.
For instance, a linear integrator can be implemented in the NSAT by just solving the equation
$x_0[t+1] = x_0[t] + (x_0[t] + \sum_{j=1}^{n}w_{ij}s_j[t])$, where the sum reflects the synaptic input to the neuron.

\subsection{Amari's Neural Fields}
Neural fields are integro-differential equations usually modeling spatiotemporal dynamics of a cortical sheet, firstly introduced by Shun'ichi Amari in $1977$ in his seminal paper \cite{Amari77_dynapatt}.
Neural fields have a rich repertoire of dynamics~\cite{Bressloff11_spatdyna} (waves, breathers, stationary solutions, winner-take-all) and are thus key components of neural computational models.
The original Amari's neural field equation is given by:
\begin{align}
	\label{eq:amari}
	\tau \frac{\Pa u(r, t)}{\Pa t} &= -u(r,t) + I_{\text{ext}} + h
    							   +\int_{\Omega} w(|r-r'|) f(u(r', t)) dr',
\end{align}
where $u(r,t)$ is the average neural activity at position $r$ and at time $t$ of a neural population, $\tau$ is a time constant, $\Omega$ denotes a compact subset of $\mathbb{R}^q$, where $q \in \mathbb{N}_{\geq 1}$,
$w(|r-r'|)$ is a connectivity function that defines the connectivity strength between neurons at positions $r$ and $r'$.
$I_{\text{ext}}$ is an external input that is applied on the neural field (subcortical inputs for instance) and $h$ is the resting potential of the neural population.
The function $f(r)$ is the activation or transfer function of the system and in Amari's case is a Heaviside function:
\begin{align}
f(r) &= \left\{
    \begin{array}{ll}
      1, \qquad \text{if } x > 0\\ 
      0,  \qquad \text{otherwise.}      \end{array}
  \right.
\end{align}
The kernel function in this case is a Difference of Gaussians (DoG, see SI Fig.~\ref{Fig:nf_amari}(a)), 
\begin{align}
	\label{eq:kernel}
	w(r) &= K_e \exp \Big(-\frac{r^2}{2\sigma_e ^2} \Big) - K_i \exp \Big( -\frac{r^2}{2 \sigma_i^2} \Big),
\end{align}
where $K_e$, $K_i$ and $\sigma_e$, $\sigma_i$ are the excitatory and inhibitory amplitudes and variances, respectively.

Here, we show the implementation of neural fields in the NSAT framework. 
First, we observe that the dynamics of each $i$ unit in Eq.~\eqref{eq:amari_d} is a leaky integrate-and-fire neuron if we consider $f(r)$ as a pre--synaptic spike--event indicator function.
Taking into account that the transfer function $f(r)$ is a Heaviside, we can then model every 
unit $i$ as a leaky integrate--and--fire neuron. 
This implies that the first state component of neuron $i$ reflects the $i$--th neural field unit, and the rest of the neuron's state components remain idle.
This methodology has been previously used to implement spiking neural fields \cite{Vazquez_etal11_visuatte,Vangel_etal15_stocasyn}.

\begin{figure}[tbph!]
	\centering
    \includegraphics[width=0.5\textwidth]{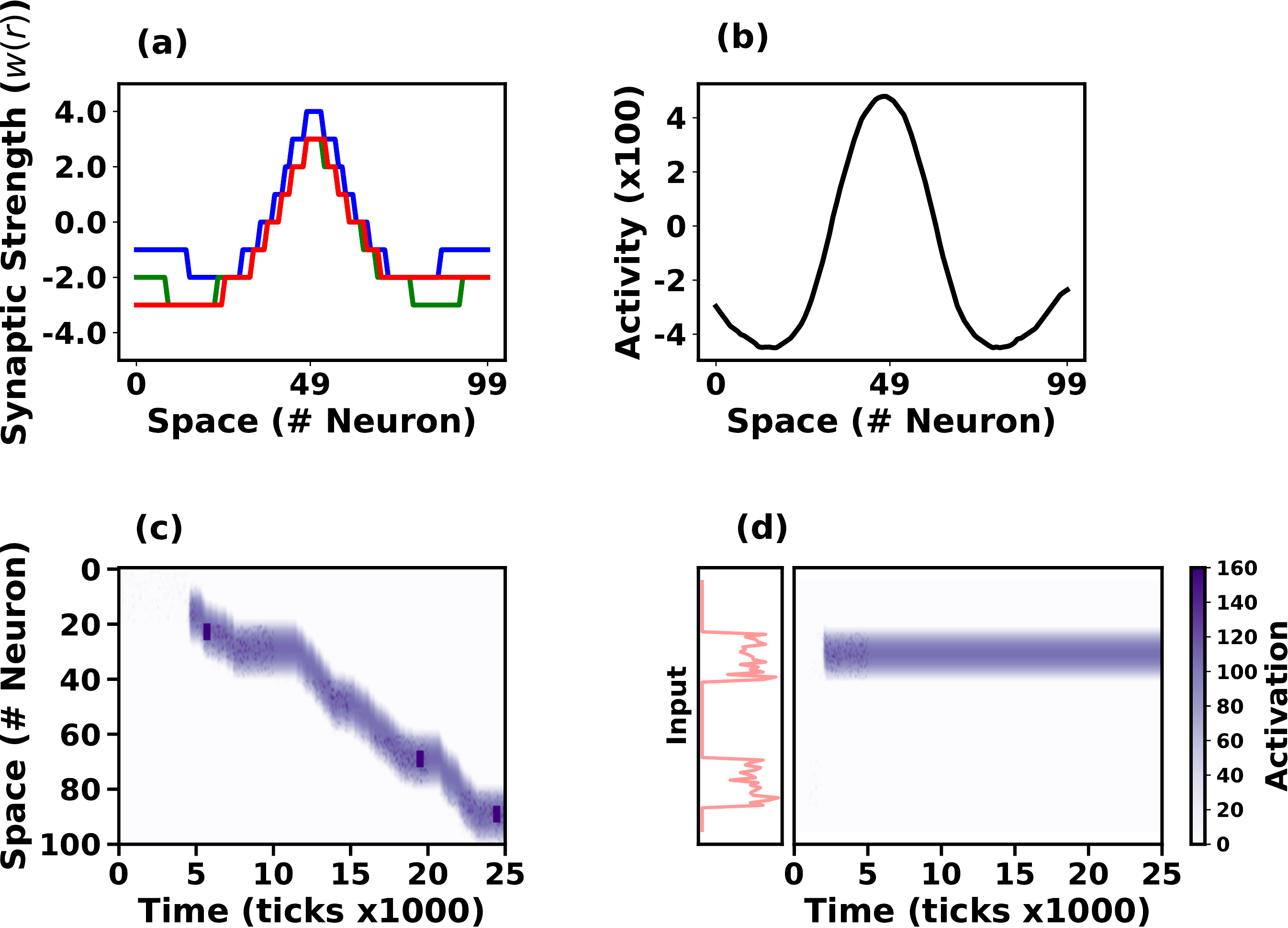}
    \caption{{\bfseries \sffamily Neural Field Implementation.} Three different neural field models were
    simulated in NSAT. 
    Three different lateral connectivity kernels ($w(r)$) are illustrated in ({\bfseries \sffamily a}). 
    Blue, green and red lines correspond to stationary solution, action--selection and tracking neural
    field models, respectively. 
    The first neural field model generates a stationary (or ``bump'') solution ({\bfseries \sffamily b}), 
    a tracking neural field model follows a moving target on the vertical axis ($y$) as ({\bfseries \sffamily c})
    illustrates. Finally the action--selection model selects one out of two input stimuli ({\bfseries \sffamily d}), red lines indicate the firing rate of the two stimuli. In panels ({\bfseries \sffamily c}) and 
    ({\bfseries \sffamily d}), the purple colormap indicates the neural activity (white--no activity,
    purple--high activity).}
    \label{Fig:nsat_nf}
\end{figure}

We quantize the kernel function $w(r)$ using a uniform quantizer $Q(r) = \Delta \cdot \lfloor \frac{r}{\Delta} + 0.5 
\rfloor$ (see Fig.~\ref{Fig:nsat_nf}(a)).
Neural resetting is disabled to match the neural fields behavior (described by Eq.~\eqref{eq:amari}
and Eq.~\eqref{eq:amari_d}): Neurons fire when they reach the firing threshold, but their states do not reset 
after spiking.

We test the NSAT neural fields implementation on three different tasks.
The first model expresses a sustained activity or stationary ``bump'' solution \cite{Amari77_dynapatt}.
We simulate $100$ internal neurons, all-to-all connected and $100$ external neurons
(no dynamics) connected with internal neurons in a one-to-one relation. 
The external neurons transmit spikes generated by a Poisson distribution with maximum firing rate
$35\, \mathrm{Hz}$ for neurons indexed from $i=40$ to $i=60$
and $10\, \mathrm{Hz}$ for the rest of the neurons.
The total duration of the input signal injection is $400$ simulation ticks and the total 
simulation time is $2500$ ticks. 
Thus the input is similar to the Gaussian function used in the continuous case (see Appendix~\ref{sec:amari}
Fig.~\ref{Fig:nf_amari}). Figure~\ref{Fig:nsat_nf}(a) shows the quantized kernel $w_{ij}$ (blue line) and Fig.~\ref{Fig:nsat_nf}(b) indicates the spatial solution of the neural field implementation at the equilibrium point. 
The solution obtained with the NSAT neural field implementation in Fig.~\ref{Fig:nsat_nf}(b) is similar to the one
in Fig.~\ref{Fig:nf_amari}(a) in Appendix~\ref{sec:amari} (red line, Amari's neural field ``bump'' solution). 

The second task involves target tracking. Asymmetric neural fields have been used for solving target 
tracking\cite{Cerda_Girau13_asymneur}. We use the same number of neurons (internal and external) and the same
simulation time as above, and modify the kernel to an asymmetric one as Fig~\ref{Fig:nsat_nf}(a)
indicates (red line). The stimulus consists of Poisson-distributed spike trains that are displaced along
the $y$--axis every $500$ ticks. 
Figure~\ref{Fig:nsat_nf}(c) illustrates the NSAT neural field to track the moving target. 
In this case, a small fraction of neurons receive Poisson--distributed spike trains at a firing rate of
$50\, \mathrm{Hz}$, while the rest of the neurons do not receive any input. 

Finally, we implemented neural fields' models of action--selection and attention  \cite{Vitay_Rougier05_usinneur}.
In this case we use the same architecture as in the previous task. 
The difference is that now we have changed the kernel function and the input.
The modified kernel function has weaker excitatory component as Fig.~\ref{Fig:nsat_nf}(a) shows (green line).
The input consists of spike trains drawn from a Poisson distribution with two localized high firing rates regions ($50\, \mathrm{Hz}$, neurons indexed from $i=20$ to $i=40$ and from $i=70$ to $i=90$) for $500$ simulation ticks (all the other internal units receive no input). 
Figure~\ref{Fig:nsat_nf}(d) shows activity when we apply the input stimulus for $500$ simulation ticks. The neural field selects almost immediately one of the two stimuli and remains there during the entire simulation (even after the stimuli removal). 

We have shown how NSAT can simulate neural fields \cite{Amari77_dynapatt,Bressloff11_spatdyna}, a sort of firing rate models. 
NSAT can thus contribute a generic framework for neuromorphic implementations of neural fields \cite{Sandamirskaya13_dynaneur} and potentially enhance them with learning features, as described in the following results.

\subsection{\textcolor{black}{Supervised Event-based Learning}}
Deep neural networks, and especially their convolutional and recurrent counterparts constitute the state-of-the-art of a wide variety of applications, and therefore a natural candidate for implementation in NSAT.
The workhorse of deep learning, the gradient descent Back Propagation (BP) rule, commonly relies on high-precision computations and the availability of symmetric weights for the backward pass.
As a result, its direct implementation on a neuromorphic substrate is challenging and thus not directly compatible with NSAT.
Recent work demonstrated an event-driven Random Back Propagation (eRBP) rule that uses a random error-modulated synaptic plasticity for learning deep representations. 
eRBP builds on the recent advances in approximate forms of the gradient BP rule \cite{Lee_etal14_targprop,Lillicrap_etal16_randsyna,Baldi_etal16_learmach} for event-based deep learning that is compatible with neuromorphic substrates, and achieves nearly identical classification accuracies compared to artificial neural network simulations on GPUs \cite{Neftci_etal17_evenranda}.

\begin{figure}[!tbph]
	\centering
    \includegraphics[width=1.0\textwidth]{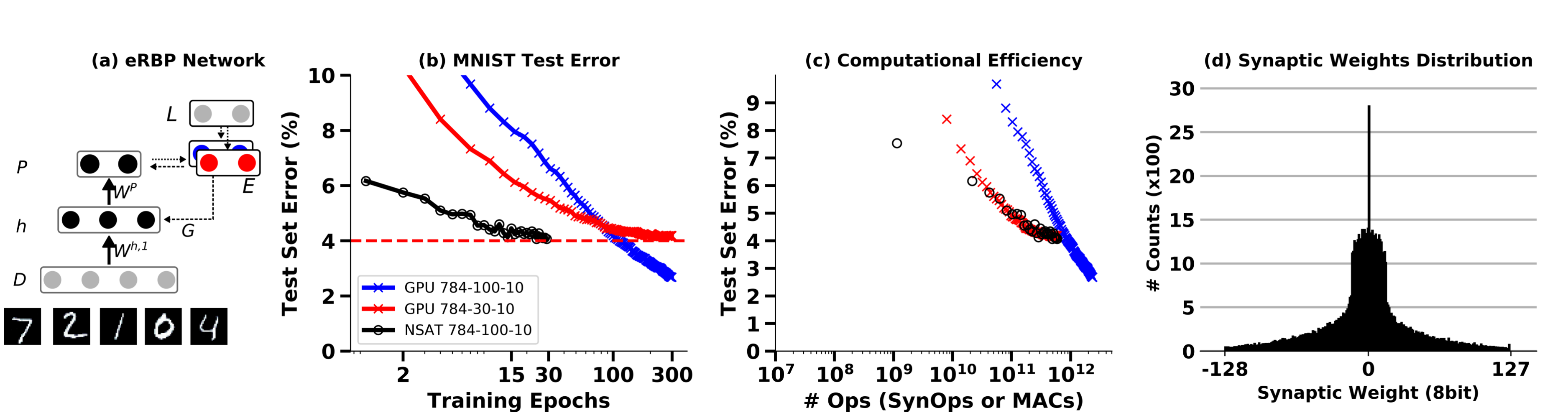}
    \caption{{\bfseries \sffamily Training an MNIST network with event-driven Random Back--propagation compared
    to GPU simulations.}
    {\bfseries \sffamily (a)} Network architecture.
    {\bfseries \sffamily (b)} MNIST Classification error on the test set using a fully connected 784-100-10
    network on NSAT (8 bit fixed--point weights, 16 bits state components) and on GPU (TensorFlow, floating--point
    32 bits). {\bfseries \sffamily (c)} Energy efficiency of learning in the NSAT (lower left is best). The number
    of operations necessary
    to reach a given accuracy is lower or equal in the spiking neural network (NSAT--SynOps) compared to the artificial
    neural network (GPU--MACs) for classification errors at or above 4\%. {\bfseries \sffamily (d)} Histogram of synaptic
    weights of the NSAT network after training. One epoch equals a full presentation of the training set.}
    \label{fig:qedp}
\end{figure}

We use a two-layer network in NSAT for eRBP equipped with stochastic synapses, and applied to learning classification in the MNIST dataset. 
The network consists of two feed-forward layers (Fig. \ref{fig:qedp}) with $N_d$ ``data'' neurons, $N_h$ hidden neurons and $N_p$ prediction (output) neurons.
The class prediction neuron and label inputs project to the error neurons with opposing sign weights.
The feedback from the error population is fed back directly to the hidden layers' neurons through random connections.
The network is composed of three types of neurons: hidden, prediction and error neurons.

The dynamics of a hidden neuron follow integrate-and-fire neuron dynamics:
\begin{subequations}
\label{eq:hidden-neurons}
\begin{align}
	\tau_{syn}\frac{\mathrm{d}V^h}{\mathrm{d}t} + V^h  &= \sum_{k} \xi(t) w_{k} s_k(t)\\
    \tau_m\frac{\mathrm{d}m^{h}}{\mathrm{d}t} + m^{h} &=  \sum_{k} g_{k}^{E} (s^{E+}_k(t) - s^{E-}_k(t))\\
    \text{if } V^h(t)>V_T & \text{ then } V_i^h\leftarrow 0 \nonumber \text{ during refractory period }\tau_{refr} \nonumber.
\end{align}
\end{subequations}
    where $s_k(t)$ are the spike trains produced by the previous layer neurons, and $\xi$ is a stochastic Bernouilli process with probability $(1-p)$ (indices $k$ are omitted for clarity).
    Each neuron is equipped with a plasticity modulation compartment $m^h$ following similar subthreshold dynamics as the membrane potential. The term $s^E(t)$ is the spike train of the error-coding neurons and $g^{E}_{k}$ is a fixed random vector drawn independently for each hidden neuron.
    The modulation compartment is not directly coupled to the membrane potential $V^h$, but indirectly through the learning dynamics.
    For every hidden neuron, $\sum_k g_{k}^{E}=0$, ensuring that the spontaneous firing rate of the error-coding neurons does not bias the learning.
    The synaptic weight dynamics follow an error-modulated and membrane-gated rule:
    \begin{equation}
          \frac{\mathrm{d}}{\mathrm{d}t} w_{j}^h \propto m^{h} \Theta(V^h)s_{j}(t).
    \end{equation}
    where $\Theta$ is a boxcar function with boundaries $b_{min}$ and $b_{max}$ and the proportionality factor is the learning rate.
    Weight values were clipped to the range $[-128,127]$ (8 bits).
    To mitigate the adverse effect of low--precision weights in gradient descent learning, we used randomized rounding where the first $r=6$ bits of $\Delta w$ were interpreted as probability.
    Prediction neurons and associated synaptic weight updates follow the same dynamics as the hidden neurons except for the modulation, where one-to-one connections with the error neurons are formed (rather than random connections).
    
Error is encoded using two neurons, one encoding positive error $E+$, the other encoding negative error $E-$. The positive error neuron dynamics are:
\begin{align}
   \label{eq:error-coding-neurons}
   \frac{\mathrm{d}}{\mathrm{d}t} V^{E+}_i &= w^{L+} (s_i^P(t) - s_i^L(t))\\
   \text{if } V^{E+}>V_T^{E} &\text{ then } V^{E+}\leftarrow V^{E+}-V_T^{E}, \nonumber
\end{align}
where $s_i^P(t)$ and $s_i^L(t)$ are spike trains from the prediction neurons and labels. 
The membrane potential is lower bounded to  $0$ to prevent negative activity to accumulate across trials.
Each error neuron has one negative counterpart neuron.
Negative error neurons follow the exact same dynamics but with $w^{L-} = -w^{L+}$. 
The firing rate of the error-coding neurons is proportional to a linear rectification of the inputs. 
For simplicity, the label spike train is regular with firing rate equal to $\tau_{refr}^{-1}$. 
When the prediction neurons classify correctly, $(s_i^P(t) - s_i^L(t)) \cong 0$, such that the error neurons remain silent.\\
Input spike trains were generated as Poisson spike trains with rate proportional to the intesenity of the pixel.
Label spikes were regular, \emph{i.e.} spikes were spaced regularly with inter--spike interval equal to the refractory period.
All states were stored in 16 bit fixed point precision (ranging from $-32768$ to $32767$), except for synaptic weights which were stored with 8 bit precision (ranging from $-128$ to $127$). 
To prevent the network from learning (spurious) transitions between digits, the synaptic weights did not update in the first $400$ ticks of each digit presentation ($1500$ ticks).

We trained fully connected feed-forward networks MNIST hand-written digits, separated in three groups, training, validation, and testing (50000, 10000, 10000 samples respectively).
During a training epoch, each of the training digits were presented in sequence during $150\mathrm{ms}$ (Fig. \ref{fig:qedp}).
Although experiments here focused on a single layer network, random back--propagation can be extended to networks with several layers, including convolutional and pooling layers \cite{Baldi_etal16_learmach}.

\textcolor{black}{Simulations of eRBP on NSAT in a quantized 784-100-10 network demonstrate results consistent with previous findings~\cite{Neftci_etal17_evenrand}. }
To highlight the potential benefits of the NSAT, we compare the number of synaptic operations (SynOp) necessary to reach a given classification accuracy (Fig. \ref{fig:qedp}(c)) to the number of multiply operations in standard artificial neural networks. 
Although larger networks trained with eRBP were reported to achieve error rates as low as $2.02\%$ \cite{Neftci_etal17_evenranda}, this NSAT network with $100$ hidden units converges to around $4\%$ error. 
As two meaningful comparisons, we used one artificial neural network with the same number of hidden units (100) and one with $30$ hidden units. 
The latter network was chosen to achieve similar peak accuracies as the simulated NSAT network.
The artificial neural network was trained using mini--batches \textcolor{black}{(the size of each mini-batch was $30$ images)} and exact gradient back--propagation using TensorFlow (GPU backend).
As previously reported, up moderate classification accuracies (here 4\%), the NSAT requires an equal or fewer number of SynOps compared to MACs to reach a given accuracy for both networks. 
We note here that only multiply operations in the matrix multiplications were
taken into account in the artificial network. Other operations such as
additions, non--linearities were ignored, which would further favor NSAT in
this comparison. \textcolor{black}{Finally, figure~\ref{fig:qedp}(d) illustrates the distribution of synaptic weights at the end of learning. It is apparent that synaptic weights concentrate mostly around $0$ with a variance of $30$. This means that $5$ bits precision is sufficient to represent the final synaptic weights. With randomized rounding enabled, lower synaptic precision during learning converges to similar results as with 8 bits of precision, but requires more time to do so (see figure~\ref{fig:5bit} in Appendix~\ref{sec:erbp}). }

These results suggests that a standard computer (\emph{e.g.} GPU) remains the architecture of choice if classification accuracy on a stationary dataset is the target, regardless of energy efficiency. 
However, the smaller or equal number of operations, compounded with the fact that a SynOp requires many fold less energy \cite{Merolla_etal14_millspik} makes a very strong argument for NSAT in terms of energy efficiency for low to moderate accuracies.
Therefore, if real--time learning is necessary, or if the streaming data is non-stationary, our results suggest that NSAT can outperform standard architectures in terms of energy efficiency \emph{at least} by a factor equal to the achieved J/MAC to J/SynOp ratio. Furthermore, the NSAT implementation of the event-driven Random Backprogation can serve as the building block for neuromorphic deep neural network architectures in the future.   

\subsubsection{Real-time learning with event-driven Random Back-Propagation}
The simplicity of the eRBP algorithm and the efficiency of cNSAT render it suitable for on-line real-time learning. 
To this end we implemented eRBP on cNSAT and interfaced it with a Davis camera \cite{Brandli_etal14_240180}.
A 28x28 pixel, center crop of the Davis camera provides spike events as input to the cNSAT while the user feeds the labels during learning through a keypad.
To interleave learning and inference, weight updates were only allowed when a label was presented. 
This mechanism was implemented within the network through an additional ``label-on'' neuron, which when active gates the positive and negative error neurons.
We trained the network using the MNIST data by alternatively presenting three different MNIST digits. The network was able to learn the MNIST classes on real-time after less than 5 presentations and the results are shown in the SI video~$1$.

\subsection{Unsupervised Representation Learning}
Synaptic Sampling Machines (S2M) are a class of neural network models that use synaptic stochasticity as a means to Monte Carlo sampling in Boltzmann machines \cite{Neftci_etal16_stocsyna}.
Learning is achieved through event--driven Contrastive Divergence (eCD), a modulated STDP rule and event--based equivalent of the original Contrastive Divergence rule \cite{Hinton02_traiprod}.
Previous work has shown that, when pre--synaptic and post--synaptic neurons firing follow Poisson statistics, eCD is equivalent to CD \cite{Neftci_etal14_evencont}.
Unsupervised learning in RBMs and S2Ms are useful for learning representations of unlabeled data, and perform approximate probabilistic inference \cite{Hinton02_traiprod,Neftci_etal14_evencont,Neftci_etal16_stocsyna}.

\begin{figure}[!tbph]
	\centering
    \includegraphics[width=1.0\textwidth]{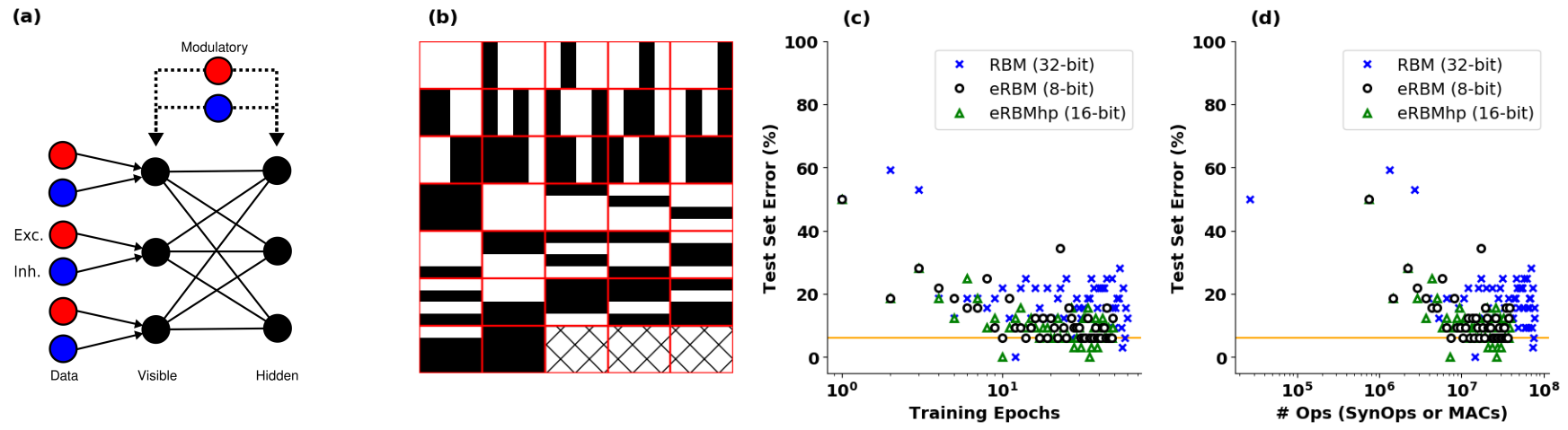}
    \caption{{\bfseries \sffamily Event--based Restricted Boltzmann Machine (eRBM).} 
    ({\bfseries \sffamily a}) Event--based RBM neural network architecture. One excitatory and one
    inhibitory unit project on the visible units providing the input to the eRBM. Two modulatory units 
    project to all the visible and hidden units. 
    ({\bfseries \sffamily b}) Bars and stripes data set.    
    ({\bfseries \sffamily c}) Training error over $300$ epochs for the eRBMhp (single precision, green triangles),
    eRBM (low precision, discs) and the classic RBM (blue crosses). The orange solid line indicates the
    minimum acceptable error for all three different algorithms.
    ({\bfseries \sffamily d}) Number of operations versus test error for the eRBMhp (single precision, green
    triangles, number of synaptic operations), eRBM (low precision, discs, number of synaptic operations)
    and the classic RBM (blue crosses, MACs). The orange dashed line indicates the minimum acceptable error for
    all three different implementations.}
    \label{fig:erbm}
\end{figure}

The NSAT synaptic plasticity dynamics are compatible with eCD under the condition that the refractory period is larger than the STDP learning window in order to maintain weight symmetry.
Here, we demonstrate on-line unsupervised learning implementing S2Ms in NSAT using the network architecture depicted in Fig.~\ref{fig:erbm}(a).
First, we use two types of input neurons, excitatory and inhibitory (red and blue nodes in Fig.~\ref{fig:erbm}(a), respectively).
These are the external units that provide the inputs to the event-based Restricted Boltzmann Machine (eRBM) visible units and their synaptic strengths are constant during learning. 
The visible units (see Fig.~\ref{fig:erbm}(a)) are all--to--all connected with the hidden ones.
Two modulatory units, one excitatory and one inhibitory, are connected with the visible and hidden units (black and gray nodes in Fig.~\ref{fig:erbm}(a)).
The two modulatory units are active in an alternating way, providing an implementation for the positive (only the excitatory unit is on) and the negative (only the inhibitory unit is active) phases of the Contrastive Divergence rule. 

In the S2M, weight updates are carried out even if a spike is dropped at the synapse.
This speeds up learning without adversely affecting the entire learning process because spikes dropped at the synapses are valid samples in the sense of the sampling process.
During the data phase, the visible units were driven with constant currents equal to the logit of the pixel intensity (bounded to the range $[10^{-5}, 0.98]$ in order to avoid infinitely large currents), plus a white noise process of low amplitude $\sigma$ to simulate sensor noise.

We run the eRBM with eCD using single precision arithmetic (eRBMhp--integers of $16$ and eRBM--$8$ bits),
as well as a classical RBM with batch size $32$ samples as a reference, on the bars and stripes data set 
\cite{MacKay03_infotheo} (see Fig.~\ref{fig:erbm}(b)).
We trained eRBM and eRBMhp using $32$ samples per epoch and for $50$ epochs. The RBM was trained using
$1$ batch of $32$ samples and $3000$ epochs (until it reaches a similar error as the eRBM and eRBMhp did).
At every epoch of the eRBM and the eRBMhp learning we run a test on all $32$ different samples and measure the classification error (how many missed classifications), whereas testing is undertaken every 50 epochs in the RBM. 
Figure \ref{fig:erbm}(c) shows the test set error against the training epochs.
The eRBM (black discs) and eRBMhp (green triangles) approach the performance of the classical RBM
(blue crosses) faster. Figure~\ref{fig:erbm}(d) shows the test set error against the number of
operations required for each of the implementations to reach the minimum acceptable error (orange solid
line).
Similarly to the supervised learning case, eRBM (black discs) and eRBMhp (green triangles) perform less or the same number of operations 
(synaptic operations) with the classical RBM (MACs). 
The three panels in Appendix~\ref{sec:rbm} Fig.~\ref{fig:erbm_rf}(a), (b) and (c) illustrate the synaptic 
weights (receptive fields) of hidden units for the RBM, eRBMhp and eRBM, respectively. 
For all three implementations we used $100$ hidden units and $18$ visible ones. 
It is apparent that the receptive fields are qualitatively similar among the three different implementations (for illustration purposes we show only $64$ out of $100$ receptive fields). 

The similarity of this S2M implementation with previous ones and the RBM suggest that NSAT is capable of 
unsupervised learning for representation learning and approximate probabilistic inference at SynOp -- MAC parity.
\textcolor{black}{This NSAT implementation of S2Ms requires symmetric (shared) connections and is thus limited to single core implementation. This requirement can be overcome with random contrastive Hebbian learning, as described in ~\cite{Detorakis_etal_2018_contrastive}. There we show that a systems of continuous non-linear differential 
equations compatible with NSAT neural dynamics is capable of representation learning similarly to restricted Boltzmann machines, while improving the speed of convergence at maintaining high generative and discriminative accuracy on standard tasks.}

\subsection{Unsupervised Learning in Spike Trains}
So far, the results have mostly focused on static data encoded in the firing rates of the neurons.
The NSAT learning rule is capable of learning to recognize patterns of spikes.
Here we demonstrate a recently proposed post-synaptic membrane potential dependent plasticity rule~\cite{Sheik_etal16_membneur} for spike train learning. 
Unlike STDP, where synaptic weights updates are computed based on spike timing of both pre-- and post--synaptic neurons, this rule triggers a weight update only on pre--synaptic spiking activity.
The neuron and synapse dynamics are governed by the following equations.
\begin{subequations}
	\begin{align}
	\tau_m \frac{\mathrm{d}V}{\mathrm{d}t} &= -V + \sum_{j=1}^{N} w_{j} s_j(t), \\
	\tau_{\mathrm{Ca}} \frac{\mathrm{dCa}_i}{\mathrm{d}t} &= -\mathrm{Ca} + w^\gamma s(t),
	\end{align}
\end{subequations}
\noindent where $V$ is the membrane potential and $\mathrm{Ca}$ is the calcium concentration, $w_j$ is synaptic weight, $\gamma$ the constant increment of the calcium concentration, and $s_j(t)$, $s(t)$ are the pre--synaptic and post--synaptic spike trains. 
The synaptic weight update dynamics are given by the equations below:
\begin{subequations}
	\begin{align}
    	\Theta_m &= \delta(V(t) > V_{lth})\eta_+ - \delta(V(t) < V_{lth})\eta_- \\
		\mathrm{mod} & = \Theta - \eta_h(\bar{\mathrm{Ca}}-\mathrm{Ca}),\\
		\Delta w_j   &= \mathrm{mod}\,  s_j(t),
    \end{align}
\end{subequations}
\noindent $V_{lth}$ is the membrane threshold that determines LTP or LTD, $\eta_+=8$ and $\eta_-=-2$ the corresponding magnitudes of LTP and LTD, $\bar{\mathrm{Ca}}$ is a constant denoting the steady-state calcium concentration and $\eta_h$ magnitude of homeostasis.

These equations can be efficiently translated to the NSAT using four ($4$) components per neuron state. Hence component $x_0$ is the membrane potential 
$V_{mem}$, $x_1$ the calcium concentration $\mathrm{Ca}_i$, $x_2$ the LTP/LTD state based on thresholded membrane component $x_0$ ($\Theta$), and
$x_3$ represents the weight modulation (value by which a weight will be updated). 
The first two state components follow exponential decay dynamics.
State components $x_2$ and $x_3$ are used to compute the effective weight updates based on the current value of membrane potential and calcium
concentration.
This is done by exploiting the fact that, at any given point in time, the weight update magnitude (if any) is given purely by the post synaptic states and is the same for every incoming spike within one time step.
\begin{figure}[!tbph]
	\centering
    \includegraphics[width=0.47\textwidth]{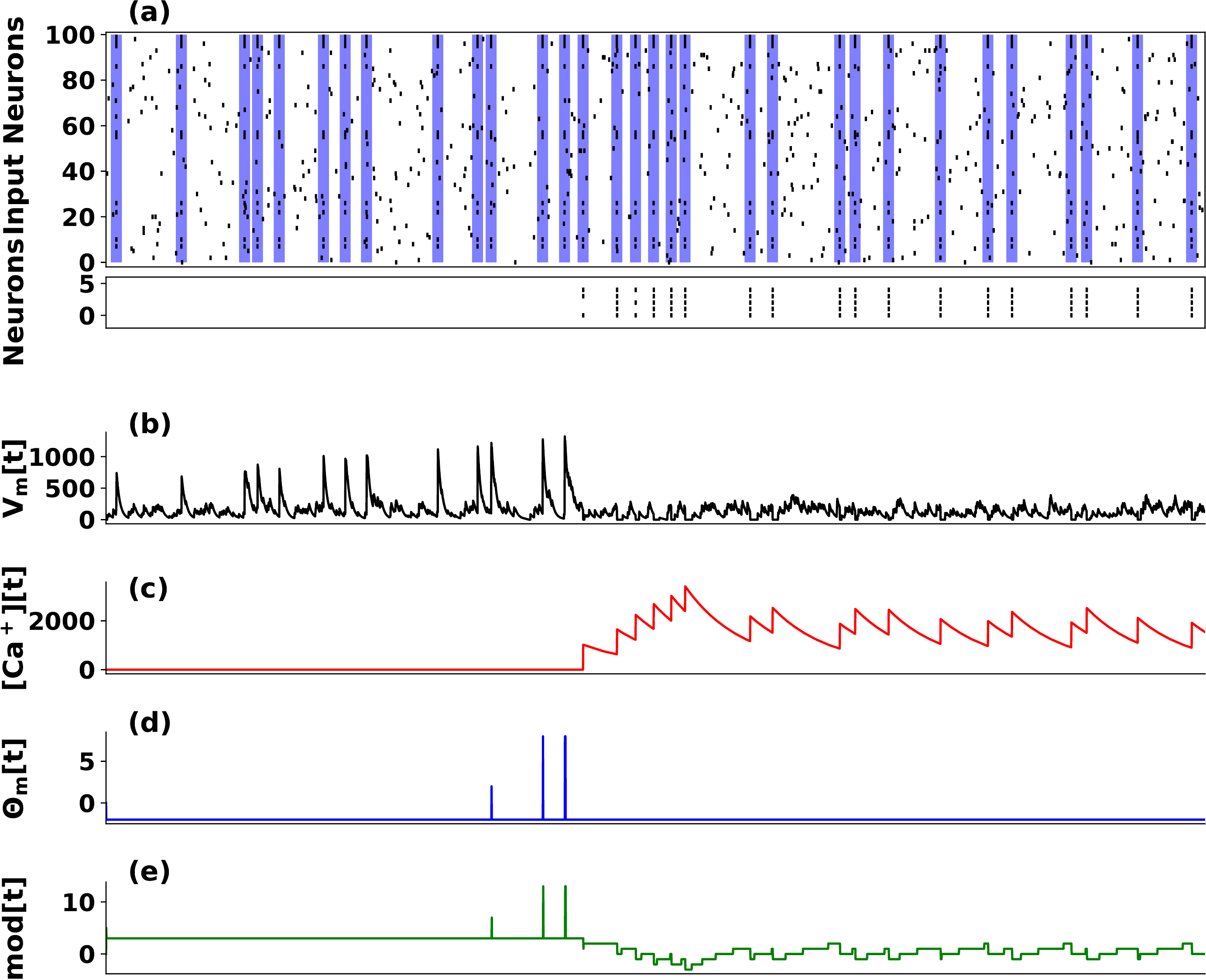}
    \caption{{\bfseries \sffamily Unsupervised Learning of Spike Patterns.}
    {\bfseries \sffamily (a)} On top is a raster plot of $100$ input spike trains (pre-synaptic neurons) 
    projecting to $5$ post-synaptic neurons. The main goal here is the post-synaptic neurons to learn 
    a hidden spike pattern indicating by the blue vertical bars. The bottom raster plot shows $5$
    post-synaptic neurons firing when they have learned the hidden spatiotemporal spike pattern. 
    {\bfseries \sffamily (b)} Indicates the membrane potential ($V_m[t]$) of the first post-synaptic 
    neuron,
    {\bfseries \sffamily (c)} its calcium concentration ($[Ca^+][t]$), 
    {\bfseries \sffamily (d)} LTP/LTD state based on thresholded membrane potential, and
    {\bfseries \sffamily (e)} its weight modulation over time.}
    \label{fig:vmem_nsat}
\end{figure}

We demonstrate these dynamics in learning to identify a hidden spike pattern embedded in noisy spike train. We use $100$ input
neurons projecting to $5$ neurons. A randomly generated fixed spike
pattern is repeatedly presented, interspersed with random spike patterns of the same firing rate as Fig.~\ref{fig:vmem_nsat}(a)
top raster plot indicates. The initial weights were randomly initialized from a uniform distribution. Over time the synaptic weights
converge such that the post--synaptic neurons (indexed $1$--$5$, black line segments in the bottom raster plot of Fig.~\ref{fig:vmem_nsat}(a))
selectively spike only on presentation of the spike pattern as Fig.~\ref{fig:vmem_nsat}(a) bottom raster plot illustrates. 
The temporal evolution of four components dynamics of the first neuron's state are given in Fig.~\ref{fig:vmem_nsat}(c)--(e).  

\textcolor{black}{This particular learning rule is a type of unsupervised temporal learning suitable for hardware implementation as demonstrated in \cite{Sheik_etal16_membneur}. When coupled with a winner-take all mechanism, it can account for the self organization of spatio-temporal receptive fields. }

\section{Discussion}
We introduced a neuromorphic computing platform and framework, called Neural and Synaptic Array 
Transceiver (NSAT), that is able to provide flexible and dynamic learning (on-line) suited for efficient digital implementation.
The NSAT takes advantage of tractable linear neural model dynamics and three-factor rules for flexibility in processing and learning.
As with existing neuromorphic systems based on Address Event Representation, only (digital) spike events are communicated across cores and between cores using event-based routing.
For reasons related to efficiency in projected digital hardware implementations, the proposed framework operates using fixed--point representation.
In addition, all the multiplications are implemented as bit shift operations, \emph{i.e} multiplications by powers of two. 
These operations are many-fold more power-efficient compared to floating-point multiply accumulates implemented in digital CMOS \cite{Horowitz14_11}. 
Taken together, the multiplier--less design, fixed--width representation and event-driven communication enable an energy-efficient and scalable system.

In this work, we demonstrated the capabilities of NSAT by showing first that neuron models with rich
behavior such as the Mihalas-Niebur neuron \cite{Mihalas_Niebur09_geneline} can be implemented, with comparable spiking dynamics \cite{Mihalas_Niebur09_geneline}.
Next, we demonstrated the simulation of neural field models \cite{Amari77_dynapatt,Bressloff11_spatdyna,Coombes05_wavebump}.
We demonstrated three core neural field behaviors, (i) a stationary ``bump'' solution (winner-take-all, working memory), (ii) an action-selection process where the neural field chooses between two input signals, and (iii) a target tracking task, where the neural field tracks a moving target.
These neural field behaviors form the backbone of many computational models using neural field, such as movement \cite{Erlhagen_Schoner02_dynafiel}, pattern generation \cite{Schoner_Kelso88_dynapatt}, soft state machines \cite{Neftci_etal13_syntcogn} and navigation \cite{Milde_etal17_obstavoi}.

NSAT is capable of on-line, event-based learning in supervised and unsupervised settings.
Embedded learning algorithms are necessary components for real-time learning, which can confer adaptability in uncontrolled environments and more fine-grained context awareness in behaving cognitive agents.
This makes NSAT suitable for environments where data are not available \emph{a priori} but are instead streamed in real--time to the device, \emph{i.e.} using event-based sensors \cite{Liu_Delbruck10_neursens}.

The implementation of machine learning algorithms in NSAT is a significant achievement, as most machine learning algorithms rely on network-wide information and batch learning. 
In contrast to machine learning algorithms implemented in standard computers, the NSAT learning is based on information that is locally available at the neuron, \emph{i.e.} (i) neurons only read weights on their own synapses, (ii) they communicate through all-or-none events (spikes), and (iii) their elementary operations are limited to highly efficient multi--compartment integrate--and--fire.
Such implementations can be more scalable compared to their Von Neumann counterparts since the access to system-wide information funnels through the von Neumann bottleneck, which dictates the fundamental limits of the computing substrate.    

Learning in NSAT is achieved using three-factor, spike-driven learning rules, \emph{i.e.} where the third factor modulates the plasticity, in addition to a programmable STDP-like learning rule \cite{Pedroni_etal16_forwtabl}.
In the NSAT, the third, modulating factor is one of the state components of the neuron.
The use of a neural state component as a third factor is justified by the fact that gradient descent learning rules in spiking neurons often mirror the dynamics of the neurons and synapses \cite{Pfister_etal06_optispik,Zenke_Ganguli17_supesupe}, while being addressable by other neurons in the network for error-driven or reward-driven learning.
Three factor rules are thus highly flexible and can support multiple different learning rules on a single neuron, thereby enabling the NSAT neuron model to be programmed for supervised, unsupervised and reinforcement learning. 
Building on previous work, we demonstrated three specific algorithms: (i) event-based deep learning and event-based Random Back-propagation algorithm for supervised settings, (ii) a Contrastive-Divergence algorithm used to train a Restricted Boltzmann Machine implemented on NSAT for unsupervised settings, and (iii) a Voltage-based learning rule for spike-based sequence learning.

The NSAT computes with limited numerical precision in its states (16 bits in this work) and in its weights (8 bits in this work). 
Often, artificial neural networks require higher precision parameters to average out noise and ambiguities in real-world data (\emph{e.g.} stochastic gradient descent) \cite{Courbariaux_etal14_lowprec,Stromatias_etal15_robuspik}, and introduce challenges at all levels
of implementation~\cite{Azghadi_etal14_spiksyna,Indiveri_Liu15_memoinfo}. 
The NSAT framework mitigates the effect of low precision using a discretized version of randomized rounding \cite{Muller_Indiveri15_rounmeth}, where a programmable number of bits are interpreted as update probability.
The randomized rounding has been demonstrated in the event-based random back propagation algorithm \cite{Neftci_etal17_evenranda}, a model that is sensitive to weight precision. Moreover, the randomized
rounding has a significant effect on the learning rate. 
We find that the networks perform well even when the synaptic weights are bounded to 256 levels (8 bits precision).

Under the selected specification, large-scale hardware implementation of NSAT is well within reach of current memory technology and can guide (and benefit from) the development of emerging memory technologies
\cite{Querlioz_etal15_bioiprog,Eryilmaz_etal16_neurarch,Mostafa_etal15_implspik,Naous_etal16_stocmode}.
While it is not possible provide direct energy comparisons at this stage, our results consistently highlight a SynOp to MAC parity in learning tasks, \emph{i.e.} the number of operations required to reach a given task proficiency. 
The significance of this parity is that the SynOp requires manyfold less energy in reported large-scale neuromorphic implementations \cite{Merolla_etal14_millspik} compared to equivalent algorithms implemented on standard computers and GPUs.
Thus, learning in NSAT is potentially more power efficient compared to standard architectures by a factor at least equal to the ratio J/MAC to J/SynOp, while achieving comparable accuracies.

\subsection{Relation to State-of-the-Art and Other Research}
Several research groups investigated brain-inspired computing as an alternative to non-von Neumann computing and as a tool for understanding the mechanisms by which the brain computes.

IBM's TrueNorth delivered impressive machine learning implementations in terms of power \cite{Esser_etal16_convnetw}.
TrueNorth's domain of application is limited to off-line learning, partly to be able to meet targeted design specifications, and partly due to the lack of suitable learning algorithms.

On-chip spike-driven, bistable learning rules were successfully demonstrated in mixed signal neuromorphic hardware.
Also, significant effort has gone into learning in digital systems \cite{Venkataramani_etal14_axnnener}: Earlier prototypes of IBM's TrueNorth \cite{Seo_etal11_45nmcmos} also demonstrated the feasibility of low-power embedded learning using STDP, and evolutionary algorithms were recently applied to FPGA-based spiking neural networks \cite{Dean_etal14_dynaadap}.
Stanford's Neurogrid team was among the first to demonstrate STDP learning in mixed-signal neuromorphic hardware \cite{Arthur_Boahen06_learsili}.
Other related neuromorphic projects on learning with neuromorphic hardware 
are the SpiNNaker \cite{Furber_etal14_spinproj} and BrainScales \cite{Schemmel_etal10_wafeneur}, as part of the Human Brain Project.
SpiNNaker is a parallel multi-core computer architecture composed of half million ARM968 processors (each core is capable of simulating $1,000$ neurons) providing a massive implementation of spiking neural networks. 
The BrainScales project and their subsequent developments are based on an analog neuromorphic chip, with its main functional blocks consisting of time-accelerated leaky integrate--and--fire neurons \cite{Aamir_etal16_hightuna}.
There, the proposed learning rule is a hybrid implementation using an on-chip SIMD processor programmable with a range of modulated STDP rules \cite{Friedmann_etal17_demohybr}.
Both projects are targeted to accelerating simulations of biological neural networks, using technologies that favor speed over compactness and/or power. 
In contrast, the NSAT design favors compactness and power over speed, and targets application-oriented, flexible, ultra low-power neural and synaptic dynamics for real-time learning tasks.

From the design perspective, the NSAT framework is closest to the TrueNorth ecosystem, but adds on-line learning capabilities and inference dynamics that are compatible with some existing event-based learning dynamics. 
For instance, NSAT allows for programmable weights and stochastic synapses, a combination that has been shown to be extremely successful in both unsupervised and supervised learning settings.

We believe that the algorithmic-driven design of the NSAT framework, combined with the provided open-source implementation will engage the research community to further investigate brain-inspired, event-based machine learning algorithms.

\subsection{NSAT Software Developments}
The software stack is a critical component to interface between the majority of potential end-users and the NSAT framework.
Our software development efforts are targeted to providing a general purpose framework for Computational Neuroscience and Machine Learning applications that combines the power of machine learning frameworks (such as TensorFlow \cite{Abadi_etal15_tenslarg}, Neon \footnote{\url{https://neon.nervanasys.com/index.html/}}) and neuromorphic hardware network description (\emph{e.g.} pyNCS \cite{Stefanini_etal14_pynckern}, TrueNorth Corelet \cite{Amir_etal13_cogncomp}).
To this end, we are expanding the software for automatic network generation in deep neural network-like scenarios (\emph{e.g.} automatic differentiation). 
Such a software stack will enable end users to 
simulate neural networks (artificial, spiking or compartmental and even firing rate models) without knowledge of NSAT's technical details.

In this article, we briefly introduced PyNSAT (see Appendix~\ref{sec:soft_implem}), an interface for our NSAT
software implementation that can serve as an Application Programming Interface (API) for the NSAT framework. 
PyNSAT offers a rapid way to program and use the NSAT framework through the Python environment,
thereby leveraging the wide capabilities of Python's application ecosystem.
Current developments of pyNSAT are targeting proof-of-concept approaches for network synthesis in machine learning applications, in-line with existing machine learning libraries such as Keras (Tensorflow) or Neon.


\section{Appendix}

\subsection{Notation and Abbreviations}

\begin{table}[htpb!]
\centering
\begin{tabular}{l|l}
Symbols & Description \\
\hline \\
${\bf A}$   	& Transition matrix \\
${\bf b}$		& Bias term \\
${\bf x}[t]$	& Neural state \\
$\pmb{\epsilon}[t]$ & STDP eligibility function \\
$K(\cdot)$ 		& STDP kernel function \\
${\bf W}$		& Synaptic weights \\
$\pmb{\theta}$	& Threshold \\
$\pmb{\eta}$	& Additive noise \\
$\pmb{\Xi}$ 	& Multiplicative noise (Bernoulli distribution) \\
${\bf X}_r$		& Reset value \\
${\bf s}[t]$	& Spike train \\
$>>$			& Right bit shift \\
$\circ$		& Hadamard product \\ 
$\diamond$		& Bit-shift multiplication operation \\
$\diamonddiamond$ 	& Zero-rounding bit shift operation \\
\end{tabular}
\label{table:notation}
\caption{{\bfseries \sffamily Notation}}
\end{table}

\begin{table}[htpb!]
\centering
\begin{tabular}{l|l}
Abbreviation & Description \\
\hline \\
NSAT		& Neural and Synaptic Transceiver Array \\
FPGA		& Field-programmable Gate Array \\
DED 		& Differed Event-driven \\
AER 		& Address Event Representation \\
AON			& Always ON \\
RLE			& Run Length Encode \\ 
STDP		& Spike-timing Dependent Plasticity \\
LTP			& Long-term Potentiation \\
LTD			& Long-term Depression \\
DSP			& Digital Signal Processing \\
RTL			& Register Transfer Level\\
SNN			& Spiking Neural Network \\
ASIC		& Application-specific integrated circuit \\
MNN			& Mihalas-Niebur Neuron \\
DoG			& Difference of Gaussians \\
SynOp		& Synaptic Operations \\
MAC			& Multiplication Accumulation \\
GPU			& Graphics Processing Unit \\ 
eCD			& event-based Contrastive Divergence \\
S2M			& Synaptic Sampling Machine \\
eRBP  		& event-based Random Back-propagation \\
RBM			& Restricted Boltzmann Machine \\
eRBM		& event-based Restricted Boltzmann Machine \\
eRBMhp		& event-based Restricted Boltzmann Machine high precision 
\end{tabular}
\label{table:abbrev}
\caption{{\bfseries \sffamily Abbreviations}}
\end{table}

\subsection{Algorithmic description of bit-shift operators}
\label{sec:algorithmic}

\begin{algorithm}[!tpbh]
	\begin{algorithmic}
   		\Function{$a \diamonddiamond x$}{} 
        \State $y = a \diamond x$
      	\If {$ y \ne 0 $ and $a=0$}
      		\State \Return $\mathrm{sign}(-y)$
        \Else
        	\State \Return $a$
        \EndIf
   		\EndFunction
	\end{algorithmic}
\caption{Zero-rounding bit shift operation ($\diamonddiamond$)}
\label{algo:round}
\end{algorithm}

\begin{algorithm}[!tpbh]
	\begin{algorithmic}[1]
    	\Function{$a \diamond x$}{}
    		\If {$ a \ge 0 $}
    			\State \Return $x \ll a$
    		\ElsIf {$ a < 0 $}
    			\State \Return $\mathrm{sign}(x)(|x| \gg -a)$
    		\EndIf
    	\EndFunction
	\end{algorithmic}
\caption{Bit shift multiplication operation ($\diamond$)}
\label{algo:diamond}
\end{algorithm}

\subsection{Software Implementation Details}
\label{sec:soft_implem}

In order to demonstrate the capabilities of the proposed NSAT framework, we implemented a multi-thread software simulator in the C programming language called cNSAT.
The software has been designed to accommodate foreseeable specifications imposed by the hardware, and thus all operations use 16-bit integer (fixed-point) and binary arithmetics (no multiplications) as described in the Difference Equations of NSAT Framework section.

\subsubsection{Data structures}
\begin{figure}[!tpbh]
	\centering
	\includegraphics[width=0.65\textwidth]{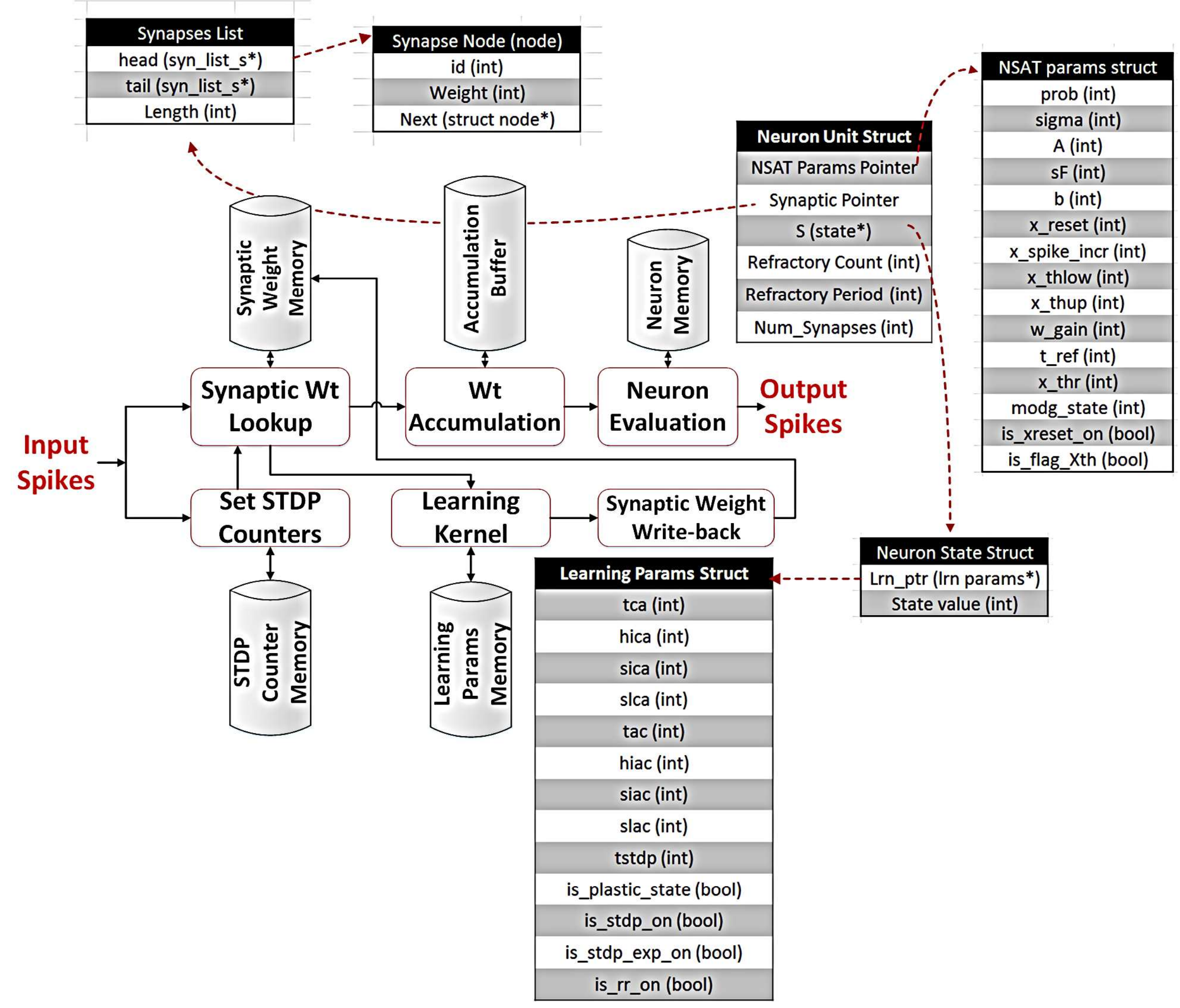}
    \caption{Schematic representation of NSAT data structures and information flow. The neuron 
    structure is the main component of the NSAT software simulator. The neuron's dynamics, learning and state parameters
    data structures are shown. Red and black arrows represent pointers and information flow, respectively. See the text
    for more information regarding the parameters and information flow.}
    \label{Fig:structs}
\end{figure}

Each thread is implemented as a large data structure that contains all the necessary data structures for implementing NSAT. 
The most significant data structure is the one that implements the neuron.
Each neuron unit structure carries all the parameters necessary for integrating the neuron's dynamics and performing learning (Fig.~\ref{Fig:structs}).
We distinguish the neurons into two main categories, external neurons and internal (NSAT) neurons.
External neurons have plastic (adjustable) post--synaptic weights and STDP counters, but no dynamics. 
Internal neuron dynamics follow Eq.~\eqref{eq:nsat_n}.
Every neuron consists of a synaptic tree implemented as a linked list containing all the post--synaptic weights and the id number of the
post--synaptic neurons.
Only internal neurons have access to the \texttt{NSAT params structure} and to the \texttt{Learning params structure}.
In addition, a state data structure is added to the internal neurons for keeping track of the dynamics (state of the neuron). 

Every thread data structure has as members the neuron's data structure (internal and external), the spike event lists, some temporary variables that are used for storing results regarding NSAT dynamics, variables that gather statistics, monitor flags, filenames strings for on--line storing of spike events or other variables (such as neuron states and synaptic weights), and finally shared to neuron parameters such as number of neurons (internal and external) within a thread, and other parameters (see Parameters paragraph below). 

\subsubsection{Random Number Generator}
The random number generator (RNG) is implemented in two different ways. First, we used the PCG RNG library
\cite{ONeill14_pcgfami} \footnote{\url{http://www.pcg-random.org/}} to implement a uniform random number generator
with long period. Based on the PCG library we implemented a Box--Muller \cite{Box_etal58_notegene} transformation in order to
acquire normal distributions for the additive noise of the NSAT framework.

The second implementation is used for simulating hardware implementations of the NSAT framework. 
In order to 
reliably generate uniformly distributed random numbers in a hardware implementation, we used a linear feedback shift register (LFSR) in combination with a cellular automata shift register (CASR) \cite{Tkacik02_hardrand}.
Such types of RNGs are suitable for hardware devices and provide a robust and reliable random number generator.
This implementation is used for bit accurate simulations of future NSAT hardware implementations. 

\subsubsection{Parameters}
NSAT framework parameters can be split into three main classes.
The first one contains global parameters related to the entire simulation and the configuration of the NSAT framework.
The second class includes parameters for neurons dynamics and for the learning process.
Figure~\ref{Fig:structs} shows the neuron's and learning parameters (\texttt{NSAT\_params} and \texttt{Learning\_params} structures, respectively).
The third class contains parameters local to each thread. 

Parameters of the first class are the number of simulation time steps (or ticks), the total number of threads, the seeds and the initial sequences for the random number generators, a flag (Boolean variable) that indicates which of the two random number generators is used, a learning flag (Boolean variable) that enables learning, and the synaptic strengths boundary control flag (Boolean variable) that enables a synaptic weight range check to more closely match hardware implementations.

The second class of parameters, the neuron parameters (refer to \texttt{NSAT params struct}), includes the state transition matrix ${\bf A}$, the constant current or bias ${\bf b}$, and ${\bf sA}$ matrix which contains the signs of matrix ${\bf A}$. ${\pmb \sigma}$ is the variance for the additive normal distributed noise and ${\bf p}$ is the blank-out probability (corresponding to ${\pmb \Xi}$ in Eq.~\eqref{eq:nsat_d}).
In addition, it contains the spike threshold (${\pmb \theta}$), the reset value (${\bf X_r}$), the upper (${\bf  X}_{\text{up}}$) and the lower boundaries (${\bf X}_{\text{low}}$) for each neural state component.
The latter two constants define the range of permitted values for each state component.
The spike increment value increases or decreases the after-spike state component value instead of resetting it.
A Boolean parameter enables or disables the reset of a neuron.
An optional parameter permits variable firing threshold, whereby component $x_1$ is used as firing threshold for the neuron.
An integer parameter defines which state component is assigned as plasticity modulator.
Finally, another parameter sets the synaptic weights gains.  

The third group consists of learning parameters.
Each neural state component has its own synaptic plasticity parameters (see Fig.~\ref{Fig:structs}), enabled by a single flag. 
The rest of the learning parameters define the STDP kernel function ($K(\cdot)$ in Eq.~\eqref{eq:learning_d}).
The STDP kernel function is either a piecewise linear function or a piecewise exponential one that can approximate the classical STDP exponential curve or other kernel functions.
The approximation uses either three linear segments for which we define the length, the height (level) and
the sign or three exponential-like segments for which we define the length, the height, the sign and the slope, thus we have eight parameters that define the kernel function (four for the causal part and four for the acausal one).

\begin{figure}[!tbph]
	\centering
    \includegraphics[width=0.45\textwidth]{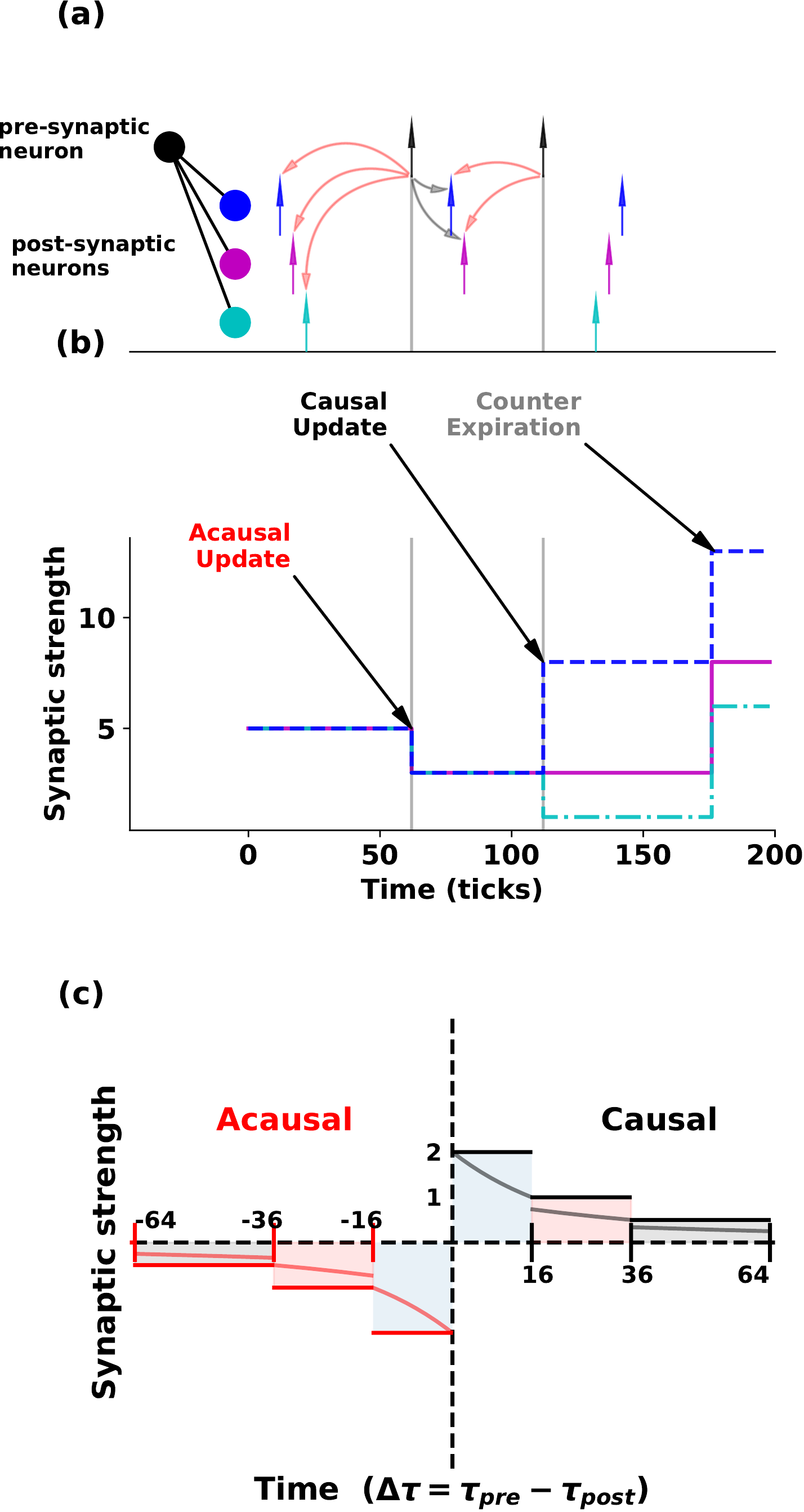}
    \caption{{\bfseries \sffamily NSAT STDP learning rule.} An actual simulation of a neuron (black dot, black
    spikes) connected to three post-synaptic ones (blue, magenta, cyan). 
    {\bfseries \sffamily (a)} A pre--synaptic neuron (black node) projects to three post-synaptic neurons (blue,
    magenta and cyan nodes). Three spikes are emitted by the post-synaptic neurons (corresponding colored arrows) 
    and then a spike is fired by the pre--synaptic neuron. Then an acausal update takes place since the post-synaptic
    spikes triggered within the acausal STDP time-window. Most recent post-synaptic spikes 
    cause a causal update within the temporal limit defined by the causal STDP time-window.  
    The light-gray lines indicate the pre-synaptic neuron's spike time, the red and black arrows illustrate
    the acausal and causal updates, respectively. 
    {\bfseries \sffamily (b)} Temporal evolution of the post-synaptic weights. The acausal and causal updates are 
    aligned with panel's (a) spikes. The gray vertical lines indicate the pre-synaptic spikes. Notice the latest 
    weights update at $187$ ticks. These causal updates are due to the expiration of the pre-synaptic neuron's 
    counter (pre-synaptic neuron does not fire a spike at that time). 
    {\bfseries \sffamily (c)} The STDP kernel function (linear and exponential approximations) used in this simulation
    (only the linear part). Black and red colors indicate the causal (positive) and the acausal (negative) parts of the 	STDP kernel function, respectively. Darker--colored lines illustrate the linear approximated STDP curves, and 
    lighter--colored ones the exponential approximation (both types are supported by the NSAT framework).}
    \label{Fig:stdp_kernel}
\end{figure}

Fig.~\ref{Fig:stdp_kernel}(c) illustrates a realization of NSAT approximated STDP kernel function. 
\texttt{tca} (\texttt{tac}) controls the length (time dimension), \texttt{hica} (\texttt{hiac}) controls the amplitude (height), \texttt{sica} (\texttt{siac}) defines the sign for the causal (acausal) part, and
\texttt{slca} (\texttt{slac}) characterizes the slope of the exponential approximation. 
On a given thread, different types of neurons and synapses can be defined by assigning them to separate parameter groups.

Finally, the third parameter group concerns the core configuration. 
Each parameter group specifies the number of internal and external units within a thread, states configurations per thread and in addition some extra parameters for the use of temporary variables necessary in simulations. 

\subsubsection{Python Interface}

In order to facilitate the use of the software simulator (and the hardware later on) we developed a high-level interface in Python.
The Python Interface (pyNSAT from now on) is based on Numpy, Matplotlib, Scipy, pyNCS and Scikit-learn Python packages. 
The pyNCS \cite{Stefanini_etal14_pynckern} is used for generating spike trains, read and write data from/to files and it provides proper tools for data analysis and visualization of simulations results. 
\begin{figure}[!tpbh]
	\centering
    \includegraphics[width=0.75\textwidth]{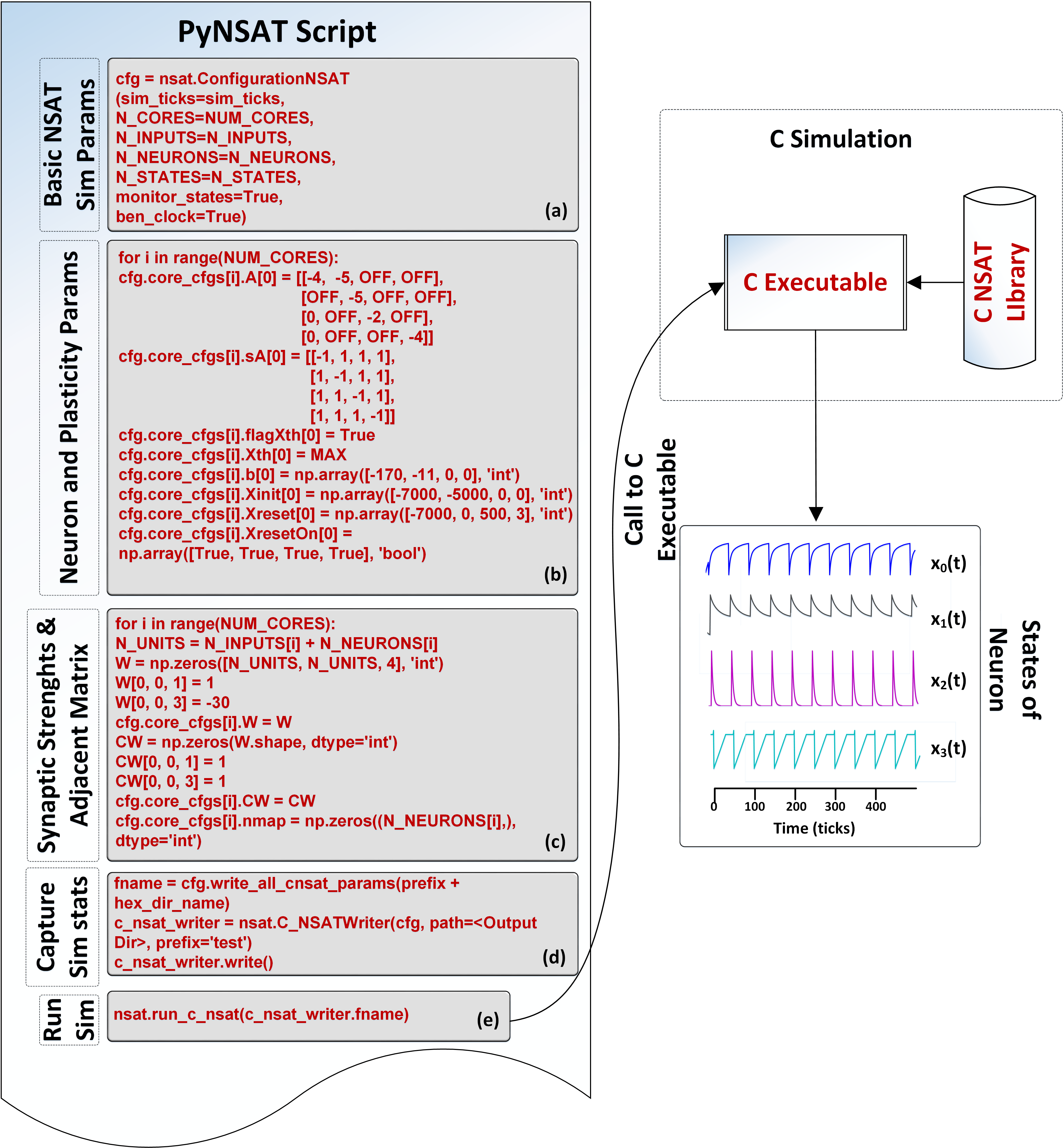}
    \caption{PyNSAT Example. Source code snippets for creating a simple simulation of a 
    single neuron with four state components ($x_i, i = 0, \ldots, 3$). The Python script
    ({\sffamily \bfseries a}) instantiates the main configuration class,
    ({\sffamily \bfseries b}) sets neural dynamics,
    ({\sffamily \bfseries c}) defines the architecture of the network (synaptic connections),
    ({\sffamily \bfseries d}) writes all the parameters files and finally
    ({\sffamily \bfseries e}) calls the C library for running the simulation. The results of the 
    simulation can be easily visualized using Python's packages.}
    \label{Fig:pynsat}
\end{figure}

Figure~\ref{Fig:pynsat} illustrates an example of pyNSAT script simulating a neuron with four state components ($x_i, i=0\ldots3$). The script mainly consists of five parts. First, we instantiate the configuration class.
This class contains all the necessary methods to configure the simulation architecture and define the global parameters for the simulation, such as the total number of threads (or cores) to be used, number of neurons per thread, number of state components per neuron per thread, number of input neurons per thread and the simulation time (in ticks), Fig.~\ref{Fig:pynsat}(a).
NSAT model parameters such as the matrix ${\bf A}$ of the NSAT dynamics, the biases ${\bf b}$ and many other parameters regarding the dynamics of the neuron and the model are defined as shown in Fig.~\ref{Fig:pynsat}(b). 
The next step is to define the synaptic connectivity (the architecture of the network), Fig.~\ref{Fig:pynsat}(c). Fig.~\ref{Fig:pynsat}(d) illustrates how the pyNSAT writer class is invoked for writing all the binary files containing the parameters that the C library will use to execute the simulation.
Finally, we call the C NSAT library and execute the simulation (see Fig.~\ref{Fig:pynsat}(e)). 

The shown example is a single neuron with adaptive threshold (more details regarding this neural model are given in Mihalas--Niebur Neuron paragraph in Results section).
After simulating the model, we can visualize the results (see the right bottom panel in Fig.~\ref{Fig:pynsat}.
The first row shows the membrane potential (state component $x_0[t]$, blue line) of the neuron, the second row indicates the adaptive threshold ($x_1[t]$, black line), and the third and fourth rows are two internal currents ($x_2[t]$ and $x_3[t]$, magenta and cyan colors), respectively. 

\subsubsection{Simulation Details}

The software simulator uses a simulator implemented in the C Programming Language.
All the source code used in this work are distributed under the GPL v$3.0$
License and are available on-line \textcolor{black}{(\url{https://github.com/nmi-lab/NSAT})}.
All simulation parameters are provided in the source code accompanying this work and in the Supplementary 
Information. 
The Python interface for the NSAT framework has been written for Python $2.7$.

\subsection{Amari's Neural Fields Simulation}
\label{sec:amari}

In order to numerically solve Eq.~\eqref{eq:amari}, we temporally discretize it using the Forward
Euler method and thus we have,
\begin{align}
	\label{eq:amari_d}
    u_i[t+1] &= u_i[t] + \frac{\D t}{\tau} \Big( -u_i[t] + I^{\text{ext}}_i + h_i 
    		 + dx\sum_{j=1}^{k} w_{ij} f(u_j[t]) \Big),
\end{align}
where $i$ is the spatial discrete node (unit or neuron), $dt$ is the Euler's method time step and $dx$ is the spatial discretization step.

\begin{figure}[!htp]
	\centering
	\includegraphics[width=0.7\textwidth]{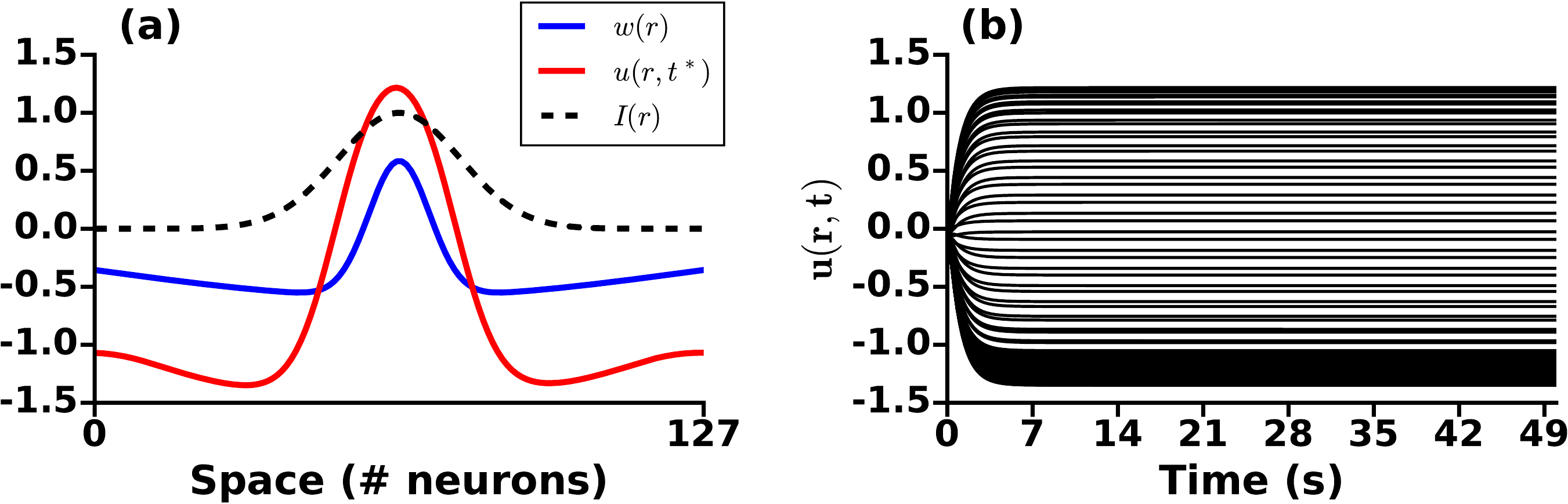}
    \caption{{\bfseries \sffamily Amari's Neural Field of Infinite Precision.}
    A numerical simulation of Eq.~\eqref{eq:amari_d}.
    In ({\bfseries \sffamily a}) Blue and red solid lines indicate the lateral connectivity kernel
    ($w(r)$) and a solution $u(r,t^*)$ for a fixed time step $t^*$, respectively. 
    The black dashed line displays the input $I_{\text{ext}}$ to the neural field
    Eq.~\eqref{eq:amari_d}. In ({\bfseries \sffamily b}) is illustrated the temporal numerical
    evolution of Eq.~\eqref{eq:amari_d} for every spatial unit $i$.
    It is apparent that after about $7$ seconds the system has reached its equilibrium point.}
    \label{Fig:nf_amari}
\end{figure}

As input to the neural field we use a Gaussian function with variance of $0.3$ (black dashed
line in Fig.~\ref{Fig:nf_amari}(a)).
At the end of the simulation the neural field has converged to its stable solution which is a ``bump'', as expected with a DoG kernel function (Fig.~\ref{Fig:nf_amari}(a), red line). 
Figure~\ref{Fig:nf_amari}(b) depicts the temporal evolution of numerical integration
of Eq.~\eqref{eq:amari} using the following parameters: $K_e = 1.5\, K_i = 0.75$, $\sigma_e = 0.1$ and
$\sigma_i = 1.0$. The integral is defined in $\Omega = [0, 1]$ and we simulate for $50$ seconds. 
After about $7$ seconds ($200$ simulation steps) the numerical solution converges
to a fixed point (see \cite{Amari77_dynapatt} for more details). 

\subsection{Supervised Event-based Learning}
\label{sec:erbp}

\begin{figure}[!htpb]
	\centering
	\includegraphics[width=0.5\textwidth]{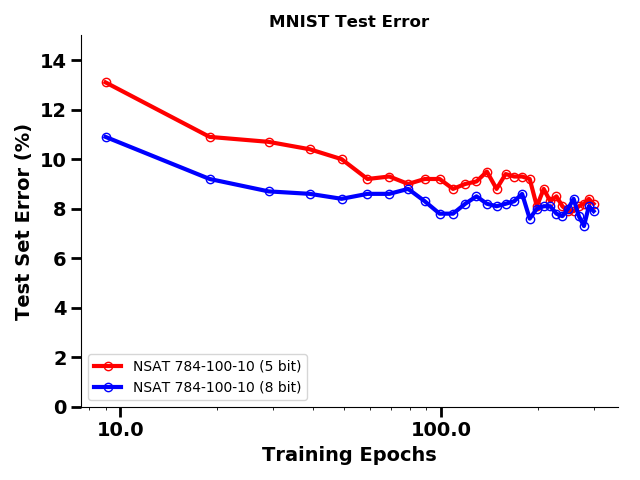}
    \caption{{\bf $8$-bit and $5$-bit synaptic precision.} The red curve
    illustrates the test error of the feed-forward network described in section
    $3.3$ trained with $5$ bit synaptic precision. The blue curve indicates the
    same network trained with $8$-bit synaptic precision. It is apparent that 
    at the initial stages of learning the higher precision ($8$-bits) leads to 
    a smaller error than the $5$-bit precision. However, at the later states of 
    learning both networks express similar behavior having similar test errors.}
    \label{fig:5bit}
\end{figure}

\subsection{Unsupervised Representation Learning}
\label{sec:rbm}

\begin{figure}[!htpb]
	\centering
    \includegraphics[width=1.\textwidth]{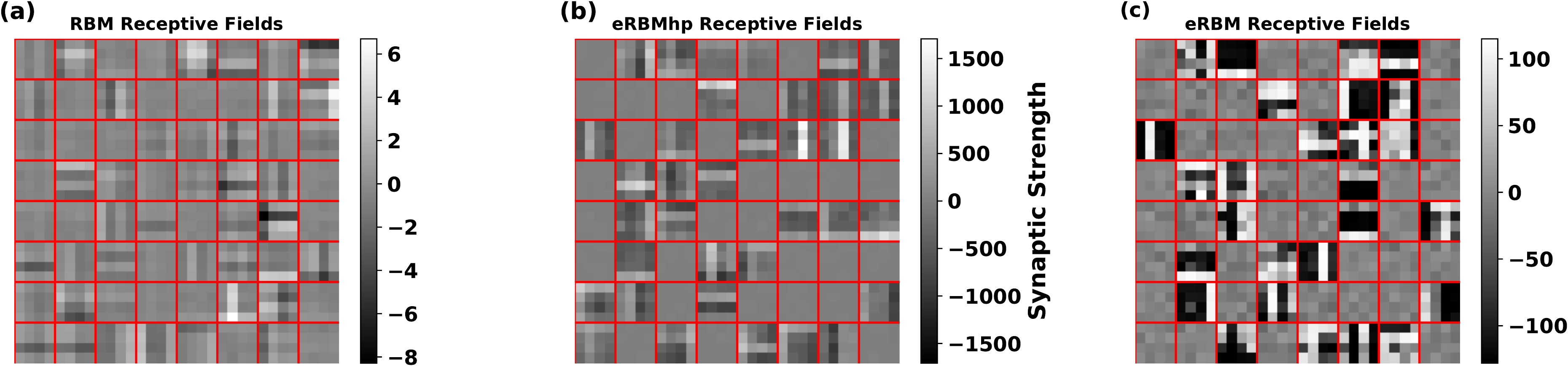}
    \caption{{\bfseries \sffamily Event-based Restricted Boltzmann Machine (eRBM) receptive fields}.
    {\bfseries \sffamily (a)} Classical RBM of infinite precision trained on the bars and stripes data set,
    {\bfseries \sffamily (b)} hRBMhp with $16$-bit precision synaptic weights, and 
    {\bfseries \sffamily (c)} eRBM with $8$-bit precision synaptic weights. } 
    \label{fig:erbm_rf}
\end{figure}

\section*{Conflict of Interest Statement}
Authors Charles Augustine and Somnath Paul were employed by company Intel Corporation.
All other authors declare no competing interests.

\section*{Author Contributions}
GD wrote the NSAT software; GD, SS and EN conceived the experiment; GD, SS and EN conducted the experiment; GD, SS and EN analyzed the results; CA and SP designed and
developed the FPGA implementation; all authors wrote and reviewed the manuscript. 

\section*{Acknowledgements}
This work was partly supported by the Intel Corporation and by the National Science Foundation under grant 1640081, and the Korean Institute of Science and Technology.

\bibliographystyle{plain}
\bibliography{biblio}

\newpage

\section*{Supplementary Material}
{\Large \bf Neural and Synaptic Array Transceiver: A Brain-Inspired
						Computing Framework for Embedded Learning}

\subsection{NSAT equations and Parameters}

In this section we provide all the transition matrices (${\bf A}$), the biases 
${\bf b}$, thresholds (${\bf \theta}$) and reset values (${\bf X}_r$) for all
simulations (see Eq.~\eqref{eq:nsat_d}) in the main article.  

\subsubsection{NSAT Neuron Dynamics Support a Wide Variety of Neural Responses}
We simulated six different neural responses (see Fig.~\ref{Fig:mnn}). For every simulation 
we used one group of NSAT parameters ($0$) and no learning at all. 
Hence parameters for Fig.~\ref{Fig:mnn}(a) are:
\begin{equation*}
\mathbf{A}^0 = \begin{bmatrix} -4 & -16 & -16 & -16  \\ 
-16 & -7 & -16 & -16  \\ 
0 & -16 & -2 & -16  \\ 
0 & -16 & -16 & -6  \\ 
\end{bmatrix} 
\quad \end{equation*} 

\begin{equation*}
\mathbf{sA}^0 = \begin{bmatrix} -1 & 1 & 1 & 1  \\ 
1 & -1 & 1 & 1  \\ 
1 & 1 & -1 & 1  \\ 
1 & 1 & 1 & -1  \\ 
\end{bmatrix} 
\quad \end{equation*} 

\begin{equation*}
\mathbf{b}^0 = \begin{bmatrix} 
-287 \\ 
-39 \\ 
0 \\ 
0 \\ 
\end{bmatrix} 
\quad \end{equation*} 

\begin{equation*}
\mathbf{Xinit}^0 = \begin{bmatrix} 
-7000 \\ 
-5000 \\ 
100 \\ 
10 \\ 
\end{bmatrix} 
\quad \end{equation*} 

\begin{equation*}
\mathbf{XresetOn}^0 = \begin{bmatrix} 
True \\ 
True \\ 
True \\ 
True \\ 
\end{bmatrix} 
\quad \end{equation*} 

\begin{equation*}
\mathbf{Xreset}^0 = \begin{bmatrix} 
-7000 \\ 
-5000 \\ 
0 \\ 
0 \\ 
\end{bmatrix} 
\quad \end{equation*} 

Parameters for Fig.~\ref{Fig:mnn}(b) are:
\begin{equation*}
\mathbf{A}^0 = \begin{bmatrix} -4 & -8 & -16 & -16  \\ 
-16 & -7 & -16 & -16  \\ 
0 & -16 & -2 & -16  \\ 
0 & -16 & -16 & -6  \\ 
\end{bmatrix} 
\quad \end{equation*} 

\begin{equation*}
\mathbf{sA}^0 = \begin{bmatrix} -1 & 1 & 1 & 1  \\ 
1 & -1 & 1 & 1  \\ 
1 & 1 & -1 & 1  \\ 
1 & 1 & 1 & -1  \\ 
\end{bmatrix} 
\quad \end{equation*} 

\begin{equation*}
\mathbf{b}^0 = \begin{bmatrix} 
-167 \\ 
-11 \\ 
0 \\ 
0 \\ 
\end{bmatrix} 
\quad \end{equation*} 

\begin{equation*}
\mathbf{Xinit}^0 = \begin{bmatrix} 
-7000 \\ 
-5000 \\ 
100 \\ 
10 \\ 
\end{bmatrix} 
\quad \end{equation*} 

\begin{equation*}
\mathbf{XresetOn}^0 = \begin{bmatrix} 
True \\ 
False \\ 
True \\ 
False \\ 
\end{bmatrix} 
\quad \end{equation*} 

\begin{equation*}
\mathbf{Xreset}^0 = \begin{bmatrix} 
-7000 \\ 
0 \\ 
500 \\ 
0 \\ 
\end{bmatrix} 
\quad \end{equation*} 

Parameters for Fig.~\ref{Fig:mnn}(c) are:
\begin{equation*}
\mathbf{A}^0 = \begin{bmatrix} -4 & -16 & -16 & -16  \\ 
-16 & -7 & -16 & -16  \\ 
0 & -16 & -2 & -16  \\ 
0 & -16 & -16 & -6  \\ 
\end{bmatrix} 
\quad \end{equation*} 

\begin{equation*}
\mathbf{sA}^0 = \begin{bmatrix} -1 & 1 & 1 & 1  \\ 
1 & -1 & 1 & 1  \\ 
1 & 1 & -1 & 1  \\ 
1 & 1 & 1 & -1  \\ 
\end{bmatrix} 
\quad \end{equation*} 

\begin{equation*}
\mathbf{b}^0 = \begin{bmatrix} 
-287 \\ 
-39 \\ 
0 \\ 
0 \\ 
\end{bmatrix} 
\quad \end{equation*} 

\begin{equation*}
\mathbf{Xinit}^0 = \begin{bmatrix} 
-7000 \\ 
-5000 \\ 
100 \\ 
10 \\ 
\end{bmatrix} 
\quad \end{equation*} 

\begin{equation*}
\mathbf{XresetOn}^0 = \begin{bmatrix} 
True \\ 
True \\ 
True \\ 
True \\ 
\end{bmatrix} 
\quad \end{equation*} 

\begin{equation*}
\mathbf{Xreset}^0 = \begin{bmatrix} 
-7000 \\ 
-5000 \\ 
0 \\ 
0 \\ 
\end{bmatrix} 
\quad \end{equation*} 

Parameters for Fig.~\ref{Fig:mnn}(d):
\begin{equation*}
\textbf{A}^0 = \begin{bmatrix} -4 & -8 & -16 & -16  \\ 
-16 & -7 & -16 & -16  \\ 
0 & -16 & -2 & -16  \\ 
0 & -16 & -16 & -6  \\ 
\end{bmatrix} 
\quad \end{equation*} 

\begin{equation*}
\mathbf{sA}^0 = \begin{bmatrix} -1 & 1 & 1 & 1  \\ 
1 & -1 & 1 & 1  \\ 
1 & 1 & -1 & 1  \\ 
1 & 1 & 1 & -1  \\ 
\end{bmatrix} 
\quad \end{equation*} 

\begin{equation*}
\mathbf{b}^0 = \begin{bmatrix} 
-194 \\ 
-11 \\ 
0 \\ 
0 \\ 
\end{bmatrix} 
\quad \end{equation*} 

\begin{equation*}
\mathbf{Xinit^0} = \begin{bmatrix} 
-3000 \\ 
-3000 \\ 
100 \\ 
10 \\ 
\end{bmatrix} 
\quad \end{equation*} 

\begin{equation*}
\mathbf{XresetOn}^0 = \begin{bmatrix} 
True \\ 
False \\ 
True \\ 
True \\ 
\end{bmatrix} 
\quad \end{equation*} 

\begin{equation*}
\mathbf{Xreset}^0 = \begin{bmatrix} 
-7000 \\ 
-6000 \\ 
0 \\ 
0 \\ 
\end{bmatrix} 
\quad \end{equation*} 

Parameters for Fig.~\ref{Fig:mnn}(e):
\begin{equation*}
\mathbf{A}^0 = \begin{bmatrix} -4 & -8 & -16 & -16  \\ 
-16 & -7 & -16 & -16  \\ 
0 & -16 & -2 & -16  \\ 
0 & -16 & -16 & -6  \\ 
\end{bmatrix} 
\quad \end{equation*} 

\begin{equation*}
\mathbf{sA}^0 = \begin{bmatrix} -1 & 1 & 1 & 1  \\ 
1 & -1 & 1 & 1  \\ 
1 & 1 & -1 & 1  \\ 
1 & 1 & 1 & -1  \\ 
\end{bmatrix} 
\quad \end{equation*} 

\begin{equation*}
\mathbf{b}^0 = \begin{bmatrix} 
-250 \\ 
-10 \\ 
0 \\ 
0 \\ 
\end{bmatrix} 
\quad \end{equation*} 

\begin{equation*}
\mathbf{Xinit}^0 = \begin{bmatrix} 
-7000 \\ 
-5000 \\ 
100 \\ 
10 \\ 
\end{bmatrix} 
\quad \end{equation*} 

\begin{equation*}
\mathbf{XresetOn}^0 = \begin{bmatrix} 
True \\ 
False \\ 
True \\ 
False \\ 
\end{bmatrix} 
\quad \end{equation*} 

\begin{equation*}
\mathbf{Xreset}^0 = \begin{bmatrix} 
-7000 \\ 
-6000 \\ 
0 \\ 
0 \\ 
\end{bmatrix} 
\quad \end{equation*} 

Parameters for Fig.~\ref{Fig:mnn}(f):
\begin{equation*}
\mathbf{A}^0 = \begin{bmatrix} -4 & -8 & -16 & -16  \\ 
-16 & -7 & -16 & -16  \\ 
0 & -16 & -2 & -16  \\ 
0 & -16 & -16 & -6  \\ 
\end{bmatrix} 
\quad \end{equation*} 

\begin{equation*}
\mathbf{sA}^0 = \begin{bmatrix} -1 & 1 & 1 & 1  \\ 
1 & -1 & 1 & 1  \\ 
1 & 1 & -1 & 1  \\ 
1 & 1 & 1 & -1  \\ 
\end{bmatrix} 
\quad \end{equation*} 

\begin{equation*}
\mathbf{b}^0 = \begin{bmatrix} 
-194 \\ 
-11 \\ 
0 \\ 
0 \\ 
\end{bmatrix} 
\quad \end{equation*} 

\begin{equation*}
\mathbf{Xinit}^0 = \begin{bmatrix} 
-7000 \\ 
-5000 \\ 
100 \\ 
10 \\ 
\end{bmatrix} 
\quad \end{equation*} 

\begin{equation*}
\mathbf{XresetOn}^0 = \begin{bmatrix} 
True \\ 
False \\ 
True \\ 
False \\ 
\end{bmatrix} 
\quad \end{equation*} 

\begin{equation*}
\mathbf{Xreset}^0 = \begin{bmatrix} 
-7000 \\ 
-6000 \\ 
1000 \\ 
0 \\ 
\end{bmatrix} 
\quad \end{equation*}

\subsubsection{Amari's Neural Fields}
In the NSAT implementation of Amari's neural fields we used only one NSAT parameters group 
($0$) and no learning parameters since there is no learning in the simulations we ran. 
Therefore, all the parameters we used for simulating the neural fields in the NSAT are
given below.

\begin{equation*}\mathbf{A}^0 = \begin{bmatrix} -2 & -16 & -16 & -16  \\ 
-16 & -16 & -16 & -16  \\ 
-16 & -16 & -16 & -16  \\ 
-16 & -16 & -16 & -16  \\ 
\end{bmatrix} 
\quad 
\mathbf{sA}^0 = \begin{bmatrix} -1 & 1 & 1 & 1  \\ 
1 & 1 & 1 & 1  \\ 
1 & 1 & 1 & 1  \\ 
1 & 1 & 1 & 1  \\ 
\end{bmatrix} 
\quad
\mathbf{b}^0 = \begin{bmatrix} 
-5 \\ 
0 \\ 
0 \\ 
0 \\ 
\end{bmatrix} 
\end{equation*} 

\begin{equation*}
\mathbf{Xinit}^0 = \begin{bmatrix} 
0 \\ 
0 \\ 
0 \\ 
0 \\ 
\end{bmatrix} 
\quad
\mathbf{XresetOn}^0 = \begin{bmatrix} 
False \\ 
False \\ 
False \\ 
False \\ 
\end{bmatrix}
\quad
\mathbf{Xreset}^0 = \begin{bmatrix} 
0 \\ 
32767 \\ 
32767 \\ 
32767 \\ 
\end{bmatrix} 
\end{equation*}

\subsubsection{Supervised Event-based Deep Learning}
Three neuron groups were used and denoted as superscripts: hidden neurons ($H$), output neurons ($O$) and error neurons ($E$).

\begin{equation*}\mathbf{A}^H = \begin{bmatrix} -7 & -16  \\ 
-16 & -6  \\ 
\end{bmatrix} 
\quad \mathbf{A}^E = \begin{bmatrix} -16 & -16  \\ 
-16 & -16  \\ 
\end{bmatrix} 
\quad \mathbf{A}^O = \begin{bmatrix} -7 & -16  \\ 
-16 & -6  \\ 
\end{bmatrix} 
\quad \end{equation*} 

\begin{equation*}\mathbf{sA}^H = \begin{bmatrix} -1 & 1  \\ 
1 & -1  \\ 
\end{bmatrix} 
\quad \mathbf{sA}^E = \begin{bmatrix} -1 & 1  \\ 
1 & -1  \\ 
\end{bmatrix} 
\quad \mathbf{sA}^O = \begin{bmatrix} -1 & -1  \\ 
-1 & -1  \\ 
\end{bmatrix} 
\quad \end{equation*}

\begin{equation*}
\mathbf{hiac} = \begin{bmatrix} 
-7 \\ 
0 \\ 
0 \\ 
\end{bmatrix} 
\end{equation*} 

\begin{equation*}\mathbf{prob\_syn}^H = \begin{bmatrix} 
9 \\ 
15 \\ 
\end{bmatrix} 
\quad \mathbf{prob\_syn}^E = \begin{bmatrix} 
15 \\ 
15 \\ 
\end{bmatrix} 
\quad \mathbf{prob\_syn}^O = \begin{bmatrix} 
9 \\ 
15 \\ 
\end{bmatrix} 
\quad \end{equation*}

\begin{equation*}\mathbf{Wgain}^H = \begin{bmatrix} 
3 \\ 
4 \\ 
\end{bmatrix} 
\quad \mathbf{Wgain}^E = \begin{bmatrix} 
4 \\ 
4 \\ 
\end{bmatrix} 
\quad \mathbf{Wgain}^O = \begin{bmatrix} 
3 \\ 
4 \\ 
\end{bmatrix} 
\quad \end{equation*} 

\begin{equation*}\mathbf{XresetOn}^H = \begin{bmatrix} 
False \\ 
False \\ 
\end{bmatrix} 
\quad \mathbf{XresetOn}^E = \begin{bmatrix} 
False \\ 
False \\ 
\end{bmatrix} 
\quad \mathbf{XresetOn}^O = \begin{bmatrix} 
False \\ 
False \\ 
\end{bmatrix} 
\quad \end{equation*} 

\begin{equation*}\mathbf{XspikeIncrVal}^H = \begin{bmatrix} 
0 \\ 
0 \\ 
\end{bmatrix} 
\quad \mathbf{XspikeIncrVal}^E = \begin{bmatrix} 
-1025 \\ 
0 \\ 
\end{bmatrix} 
\quad \mathbf{XspikeIncrVal}^O = \begin{bmatrix} 
0 \\ 
0 \\ 
\end{bmatrix} 
\quad \end{equation*} 

\begin{equation*}\mathbf{Xthlo}^H = \begin{bmatrix} 
-32767 \\ 
-32767 \\ 
\end{bmatrix} 
\quad \mathbf{Xthlo}^E = \begin{bmatrix} 
0 \\ 
0 \\ 
\end{bmatrix} 
\quad \mathbf{Xthlo}^O = \begin{bmatrix} 
-32767 \\ 
-32767 \\ 
\end{bmatrix} 
\quad \end{equation*}

\subsubsection{Unsupervised Representation Learning}

For the NSAT implementation of the eRBM and eRBMhp we used two different NSAT parameters 
groups and two different learning parameters groups, one for the visible units (V) and 
another for the hidden ones. All the parameters are given in the
following equations.

\begin{equation*}\mathbf{A}^V = \begin{bmatrix} -3 & -16 & -16 & -16  \\ 
8 & -5 & -16 & -16  \\ 
-16 & -16 & -16 & -16  \\ 
8 & -16 & -16 & -5  \\ 
\end{bmatrix} 
\quad \mathbf{A}^H = \begin{bmatrix} -3 & -16 & -16 & -16  \\ 
8 & -5 & -16 & -16  \\ 
-16 & -16 & -16 & -16  \\ 
8 & -16 & -16 & -5  \\ 
\end{bmatrix} 
\quad \end{equation*} 

\begin{equation*}\mathbf{sA}^V = \begin{bmatrix} -1 & 1 & 1 & 1  \\ 
1 & -1 & 1 & 1  \\ 
1 & 1 & -1 & 1  \\ 
1 & 1 & 1 & -1  \\ 
\end{bmatrix} 
\quad \mathbf{sA}^H = \begin{bmatrix} -1 & 1 & 1 & 1  \\ 
1 & -1 & 1 & 1  \\ 
1 & 1 & -1 & 1  \\ 
1 & 1 & 1 & -1  \\ 
\end{bmatrix} 
\quad \end{equation*} 

\begin{equation*}\mathbf{b}^V = \begin{bmatrix} 
-6000 \\ 
0 \\ 
0 \\ 
0 \\ 
\end{bmatrix} 
\quad \mathbf{b}^H = \begin{bmatrix} 
-9500 \\ 
0 \\ 
0 \\ 
0 \\ 
\end{bmatrix} 
\quad \end{equation*} 

\begin{equation*}\mathbf{hiac}^V = \begin{bmatrix} 
0 \\ 
0 \\ 
0 \\ 
\end{bmatrix} 
\quad \mathbf{hiac}^H = \begin{bmatrix} 
1 \\ 
0 \\ 
-1 \\ 
\end{bmatrix} 
\quad \end{equation*} 

\begin{equation*}\mathbf{hica}^V = \begin{bmatrix} 
0 \\ 
0 \\ 
0 \\ 
\end{bmatrix} 
\quad \mathbf{hica}^H = \begin{bmatrix} 
1 \\ 
0 \\ 
-1 \\ 
\end{bmatrix} 
\quad \end{equation*} 

\begin{equation*}\mathbf{probsyn}^V = \begin{bmatrix} 
15 \\ 
7 \\ 
15 \\ 
15 \\ 
\end{bmatrix} 
\quad \mathbf{probsyn}^H = \begin{bmatrix} 
15 \\ 
7 \\ 
15 \\ 
15 \\ 
\end{bmatrix} 
\quad \end{equation*} 

\begin{equation*}\mathbf{siac}^V = \begin{bmatrix} 
1 \\ 
1 \\ 
1 \\ 
\end{bmatrix} 
\quad \mathbf{siac}^H = \begin{bmatrix} 
-1 \\ 
-1 \\ 
-1 \\ 
\end{bmatrix} 
\quad \end{equation*} 

\begin{equation*}\mathbf{tac}^V = \begin{bmatrix} 
-35 \\ 
-35 \\ 
\end{bmatrix} 
\quad \mathbf{tac}^H = \begin{bmatrix} 
-16 \\ 
-36 \\ 
\end{bmatrix} 
\quad \end{equation*} 

\begin{equation*}\mathbf{tca}^V = \begin{bmatrix} 
35 \\ 
35 \\ 
\end{bmatrix} 
\quad \mathbf{tca}^H = \begin{bmatrix} 
16 \\ 
36 \\ 
\end{bmatrix} 
\quad \end{equation*} 

\begin{equation*}\mathbf{Wgain}^V = \begin{bmatrix} 
2 \\ 
1 \\ 
0 \\ 
0 \\ 
\end{bmatrix} 
\quad \mathbf{Wgain}^H = \begin{bmatrix} 
2 \\ 
1 \\ 
0 \\ 
5 \\ 
\end{bmatrix} 
\quad \end{equation*} 

\begin{equation*}\mathbf{Xreset}^V = \begin{bmatrix} 
0 \\ 
32767 \\ 
32767 \\ 
32767 \\ 
\end{bmatrix} 
\quad \mathbf{Xreset}^H = \begin{bmatrix} 
0 \\ 
32767 \\ 
32767 \\ 
32767 \\ 
\end{bmatrix} 
\quad \end{equation*} 

\begin{equation*}\mathbf{Xthlo}^V = \begin{bmatrix} 
-32768 \\ 
-32768 \\ 
-1 \\ 
-32768 \\ 
\end{bmatrix} 
\quad \mathbf{Xthlo}^H = \begin{bmatrix} 
-32768 \\ 
-32768 \\ 
-1 \\ 
-32768 \\ 
\end{bmatrix} 
\quad \end{equation*} 

\begin{equation*}\mathbf{Xthup}^V = \begin{bmatrix} 
32767 \\ 
32767 \\ 
1 \\ 
32767 \\ 
\end{bmatrix} 
\quad \mathbf{Xthup}^H = \begin{bmatrix} 
32767 \\ 
32767 \\ 
1 \\ 
32767 \\ 
\end{bmatrix} 
\quad \end{equation*}

\subsubsection{Unsupervised Learning in Spike Trains}

The parameters $b_2=-V_{lth} = -1216$, $b_3 = \eta_h = 5$, $w^\gamma=1024$, 
$a_{31} = -\eta_h/\bar{\mathrm{Ca}} = -9$ and $a_{00}=-4$, $a_{11}=-8$ are chosen to appropriately replicate the corresponding state decay time constants and couplings.

\begin{equation*}{\bf A}^0 = \begin{bmatrix}
-4 & 0 & -16 & -16   \\ 
-16 & 0 & -16 & 0   \\ 
-16 & -16 & -8 & -9  \\ 
-16 & -16 & -16 & 0   \\ 
\end{bmatrix} 
\quad \end{equation*} 

\begin{equation*}{\bf sA}^0 = \begin{bmatrix} 
-1 & 1 & 1 & 1   \\ 
1 & -1 & 1 & 1   \\ 
1 & 1 & -1 & -1   \\ 
1 & 1 & 1 & -1   \\ 
\end{bmatrix} 
\quad \end{equation*} 

\begin{equation*}
\mathbf{b}^0 = \begin{bmatrix} 
0 \\ 
-1216 \\ 
0 \\ 
5 \\ 
\end{bmatrix} 
\quad \end{equation*} 

\begin{equation*}
\mathbf{hiac}^0 = \begin{bmatrix} 
1 \\ 
4 \\ 
0 \\ 
\end{bmatrix} 
\quad \end{equation*} 

\begin{equation*}
\mathbf{hica^0} = \begin{bmatrix} 
2 \\ 
0 \\ 
-2 \\ 
\end{bmatrix} 
\quad \end{equation*} 

\begin{equation*}
\mathbf{tac}^0 = \begin{bmatrix} 
16 \\ 
36 \\ 
\end{bmatrix} 
\quad \end{equation*} 

\begin{equation*}
\mathbf{XspikeIncrVal}^0 = \begin{bmatrix} 
0 \\ 
0 \\ 
1024 \\ 
0 \\ 
\end{bmatrix} 
\quad \end{equation*} 

\begin{equation*}\mathbf{Xreset}^0 = \begin{bmatrix} 
0 \\ 
32767 \\ 
32767 \\ 
32767 \\ 
\end{bmatrix} 
\quad \end{equation*} 

\begin{equation*}
\mathbf{Xthlo}^0 = \begin{bmatrix} 
0 \\ 
-2 \\ 
-32767 \\ 
-32767 \\ 
\end{bmatrix} 
\quad \end{equation*} 

\begin{equation*}
\mathbf{Xthup}^0 = \begin{bmatrix} 
32767 \\ 
8 \\ 
32767 \\ 
32767 \\ 
\end{bmatrix} 
\quad \end{equation*}

\end{document}